\documentclass[twoside]{article}

\usepackage[letterpaper,margin=1in]{geometry}
\usepackage{authblk}

\newcommand{\aistatstitle}[1]{\begin{center}{\Large\bfseries #1\par}\end{center}\vspace{1em}}
\usepackage[round]{natbib}
\renewcommand{\bibname}{References}
\renewcommand{\bibsection}{\subsubsection*{\bibname}}

\usepackage{microtype}
\usepackage{graphicx}
\usepackage{subcaption}
\usepackage{booktabs} % for professional tables

\usepackage{multirow}

\usepackage{hyperref}

\usepackage{algorithm}
\usepackage{algpseudocode}   % from the algorithmicx bundle

\usepackage{amsmath}
\usepackage{amssymb}
\usepackage{mathtools}
\usepackage{amsthm}
\usepackage{wrapfig}

\usepackage{enumitem}

\usepackage[capitalize,noabbrev]{cleveref}

\theoremstyle{plain}
\newtheorem{theorem}{Theorem}[section]
\newtheorem{proposition}[theorem]{Proposition}
\newtheorem{lemma}[theorem]{Lemma}

\theoremstyle{definition}
\newtheorem{definition}[theorem]{Definition}

\theoremstyle{remark}
\newtheorem{remark}[theorem]{Remark}

\usepackage{listings}

\usepackage[textsize=tiny]{todonotes}

\usepackage{etoolbox}

\newtoggle{anonymize}
\toggletrue{anonymize} % Set this to true or false to toggle
\newcommand{\stexnintlink}[1]{} % this is a special link for stexn
\newcommand{\stexndeletedintlink}[1]{} % this is deleted internal link
\newcommand{\stexnextlink}[1]{} % just a glorified comment
\newcommand{\stexndeletedextlink}[1]{} % mark link for deletion

\newcommand{\stexntodo}[1]{} % glorified comment
\newcommand{\stexnDELETEDtodo}[1]{} % marked for deletion

\newcommand{\bbR}{\mathbb{R}}

\newcommand{\bfX}{\mathbf{X}}

\newcommand{\bfz}{\mathbf{z}}

\newcommand{\calL}{\mathcal{L}}

\newcommand{\calO}{\mathcal{O}}

\newcommand{\kappabar}{\bar{\kappa}}

\newcommand{\redact}[2]{#2}

\usepackage{xspace}
\def\shancs{\textsc{shancs}\xspace}
\begin{document}



\title{A Fast Algorithm for the Familywise Loss\thanks{Code: \url{https://github.com/Leelavathi-R/fast-familywise-loss}}}
\author[1]{Leelavathi Raja}
\author[1]{Amirreza Eshraghi}
\author[2]{Vinay Prabhakar}
\author[3]{Mark Blacher}
\author[4]{Han Bao}
\author[1]{Yutong Wang}
\affil[1]{Illinois Institute of Technology}
\affil[2]{Modular}
\affil[3]{Friedrich Schiller University Jena}
\affil[4]{The Institute of Statistical Mathematics \& Tohoku University \& RIKEN AIP}
\date{}
\maketitle

\begin{abstract}
  The familywise (FW) loss for multiclass classification,  introduced by Duchi, Ruan, and Khosravi, achieves universal equivalence to the 0-1 loss, and thus has stronger theoretical properties than cross entropy (CE).  Prior work, however, left open whether FW is as computationally efficient as CE.  First, on the theoretical side, we prove that the FW gradient can be computed in $O(K)$ time, where $K$ is the number of classes.  Second, on the practical side, we develop optimized kernels for computing the forward and backward passes of FW which outperform CE in terms of speed on both CPU and GPU. 
\end{abstract}

\section{Introduction}

Loss function is a critical component of deep learning models. Back-propagation starts from taking the gradient of the loss function.
Therefore, the gradient of the loss influences updates for all the model parameters.
Yet the choice of cross entropy has largely remained unchanged in classification.

The cross entropy (CE) is deeply rooted in statistics, appeared as a key ingredient of the negative log likehood in multinomial logistic regression \citep{hastie2009elements}.
Moreover, cross entropy enjoys the desirable classification-calibration property \citep{tewari2007consistency,zhang2004statistical}.

However, CE is \emph{not} the only loss with this property.
Recently, \cite{duchi2018multiclass}
shows that the \emph{familywise (FW) loss}
is superior to the CE from a classification-calibration point of view.
From the practical side, \cite{fathony2016adversarial} shows that the FW loss is effective for training linear classifiers compared to other multiclass hinge losses.

For $K$-class classification, the familywise loss is defined as
\begin{align}
  \label{equation:preview:definition-of-familywise-loss}
  \calL(\bfz,y) &= 1 - z_{y} +
          {
          \max_{\kappa = 1,\dots, K} \left\{
          \tfrac{1}{\kappa} (
          \textstyle\sum_{j =1}^{\kappa} z_{(j)}) - \tfrac{1}{\kappa}
          \right  \}
        },\\ &\qquad \mbox{where} \quad
        z_{(1)} \ge z_{(2)} \ge \cdots \ge z_{(K)}. \notag
\end{align}
Here, the \(\bfz = (z_{j})_{j=1}^K\) are the (pre-softmax) logits
 of a model such as a neural network.
Prior works leave open the question of computationally efficient implementation of the familywise loss.
Cross entropy and its gradient function can be computed in \(O(K)\) time, where \(K\) is the number of classes. On the other hand, naive implementation of the FW loss is in \(O(K \log K)\) time, due to the sorting operation in \eqref{equation:preview:definition-of-familywise-loss}.

In this work, we show that the FW loss and its subgradient function can be computed in \(O(K)\) as well (see Theorem~\ref{theorem:linear-time-computation-of-the-gradient} below).
Let \(\bar{\kappa}\) be the argmax in \Cref{equation:preview:definition-of-familywise-loss}. The main observation is that we need only the top \(\bar{\kappa}\) entries \(z_{(1)}, \dots, z_{(\bar{\kappa})}\) and \(z_{y}\), which affect the gradient of
\(\calL(\bfz, y)\).
In fact, this gradient
can be calculated as
\begin{align}
  \label{equation:gradient-of-the-familywise-loss}
  \nabla_{\bfz} \calL(\bfz, y) &= -\mbox{one-hot@\(y\)}
  \\ &\quad +
  (1/ \bar{\kappa}) \cdot \mbox{many-hot@\(\{(1),\dots, (\bar{\kappa})\}\)}. \notag
\end{align}
See \Cref{remark:prove-for-the-gradient-formula} below for mathematical details.
Thus, the majority of computational work is estimating \(\bar{\kappa}\) and retrieving the top-\(\bar{\kappa}\) logits \(z_{(1)} ,\dots, z_{{(\bar{\kappa})}}\).

\textbf{Our contributions}. We advance both the theory and application of the familywise  along the following directions:

\emph{Theory.}
We show that the familywise loss and a subgradient are computable in \(\calO(K)\) time in the worst
case, where \(K\) is the number of classes, improving on the \(\calO(K \log K)\) sorting-based
computation (\Cref{theorem:linear-time-computation-of-the-gradient}).
The proof rests on an identity showing that the ``max'' in
\Cref{equation:preview:definition-of-familywise-loss} equals the root of a monotone function of one
variable (\Cref{theorem:threshold-identity}).
We also prove a \emph{bitonicity property} of the terms inside the ``max''
(\Cref{proposition:bitonicity-property-of-normalized-cumulative-sums}).

\emph{Efficient CPU and CUDA implementations.}
Building on this theory, we implement custom kernels for the loss and its gradient on multicore
CPUs and GPUs.
Our CPU implementation is faster than PyTorch cross entropy (\Cref{fig:CPU-time-comparison}).
Our CUDA kernel is faster than PyTorch cross entropy, naive (sorting-based) FW and top-\(k\)-based
FW (\Cref{fig:GPU-time-comparison}).

\section{Related work}

\begin{figure*}[t]
  \centering
\includegraphics[width=\textwidth]{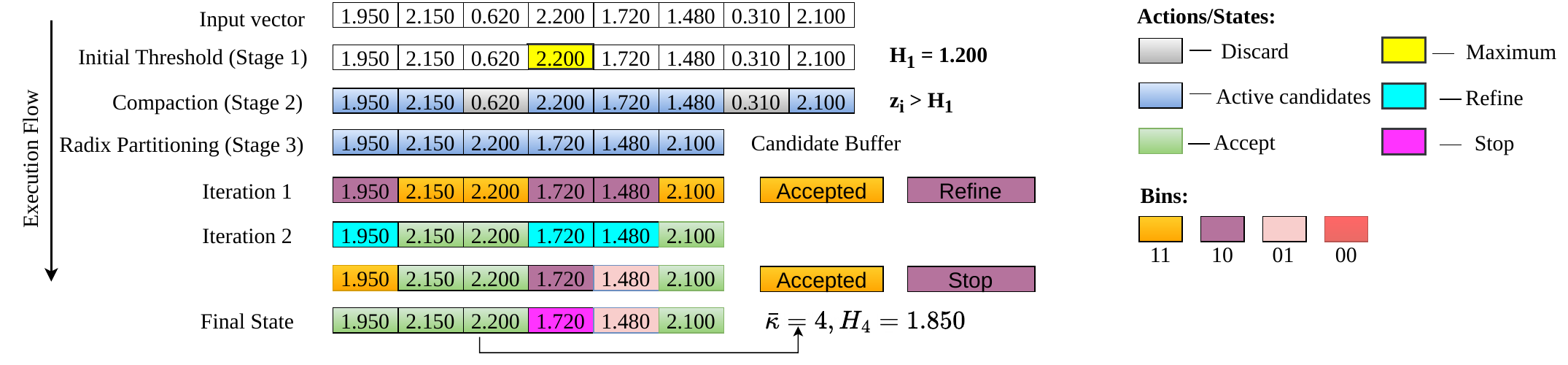}
\caption{
An illustrative example (per-sample) of \Cref{alg:radix-partitioning}. The input vector consists of eight logits, with $r=2$ bits processed per pass, resulting in $2^r=4$ bins represented by different colors. After determining the initial threshold (Stage 1), the remaining logits are copied to the candidate buffer (Stage 2) and then partitioned into radix bins (Stage 3). The bins are successively accepted or refined until the stopping condition is reached, yielding the final maximizing $\kappabar$.
}
\label{fig:gpu-running-example}
\end{figure*}

\textbf{Previous work on the FW loss}.
While less widely known, the FW loss appeared in \citep{bartlett2008classification}, however only in the binary classification case.
Later, \cite{duchi2018multiclass} defined this loss for multiclass classification where \(K \ge 2\) and proved the \emph{universal equivalence to 01 loss} property underlying the advantage of FW loss over the CE loss. Below, we will say ``universal equivalence'' to mean ``universal equivalence to 01 loss''.

Roughly speaking, the universal equivalence \cite{duchi2018multiclass} says the following:
Let  \(\mathcal{M}\) be the set  of all {measurable functions} on the sample space, and \(\mathcal{Q}\) be either \(\mathcal{M}\) itself or a finite subset of \(\mathcal{M}\). Let \(\{f_n\}_{n=1,2,\dots}\) be a sequence of functions in \(\mathcal{Q}\).
Whenever the testing \(\calL\)-risk converges to the Bayes-risk (i.e., the best theoretically achievable \(\calL\)-risk)
  \(    \mathbb{E}_{\bfX,Y} [ \calL(f_{n}(\bfX),Y)  ]
    \,\,
    \to
    \,\,
    \inf_{f \in \mathcal{Q}}
    \mathbb{E}_{\bfX,Y} [ \calL(f(\bfX),Y)  ]
    \,\, \mbox{ as \(n \to \infty\)}
\)
  the testing \emph{01 loss} \emph{also converges} to the theoretical best-case scenario
  \(    \mathbb{E}_{\bfX,Y} [ \mathrm{argmax}(f_{n}(\bfX)) \ne Y)  ]
    \,\,
    \to
    \,\,
    \inf_{f \in \mathcal{Q}}
    \mathbb{E}_{\bfX,Y} [ \mathrm{argmax}(f_{n}(\bfX)) \ne Y)]
        \,\, \mbox{ as \(n \to \infty\)}.
\)

  This is in contrast to a more widely known definition of \emph{classification-calibration}, which only applies when \(\mathcal{Q} = \mathcal{M}\).
  Thus, universal equivalence is more applicable than classification-calibration.
  The fact that the FW loss has the universal equivalence property was proved in \citet[Examples 7 \& 8]{duchi2018multiclass}.
  On the other hand, the CE loss has only the classification-calibration property.

  In summary, the FW loss \emph{should}, in theory, outperform the CE loss. However, this does not necessarily mean that FW loss is computationally efficient, nor that theoretical advantage carries over to real world tasks. In this work, we fill both gaps.

  \textbf{Classification-calibration}.
  Establishing the desirable property of classification-calibration for binary loss functions was initiated in \cite{bartlett2006convexity} and refined subsequently by many authors. See \cite{frongillo2021surrogate} and the references there-in. On the multiclass classification side, the work was initiated by \cite{tewari2007consistency,zhang2004statistical}.

  The theory of classification-calibration played an important role in designing and comparing multiclass loss functions in the context of multiclass linear models. See \cite{fathony2016adversarial,dogan2016unified} for recent works on the impact of the theory on practice.

Recently, several works have studied loss functions beyond cross entropy, such as the square loss
and ``squentropy'' \citep{hui2020evaluation,hui2023cut} and losses robust to label noise
\citep{zhang2018generalized,amid2019robust}.
Among these, the familywise loss of \cite{duchi2018multiclass} has the strongest theoretical
guarantees. We make it fast: we prove that it can be computed in \(\calO(K)\) time and give CPU
and GPU kernels that are faster than cross entropy.
With off-the-shelf automatic learning-rate schedulers
\citep{chandra2022gradient,ivgi2023dog,mishchenko2024prodigy} it is competitive with cross entropy
in accuracy (\Cref{section:appendix:experiments}).

\begin{figure*}[t]
  \centering
\includegraphics[width=\textwidth]{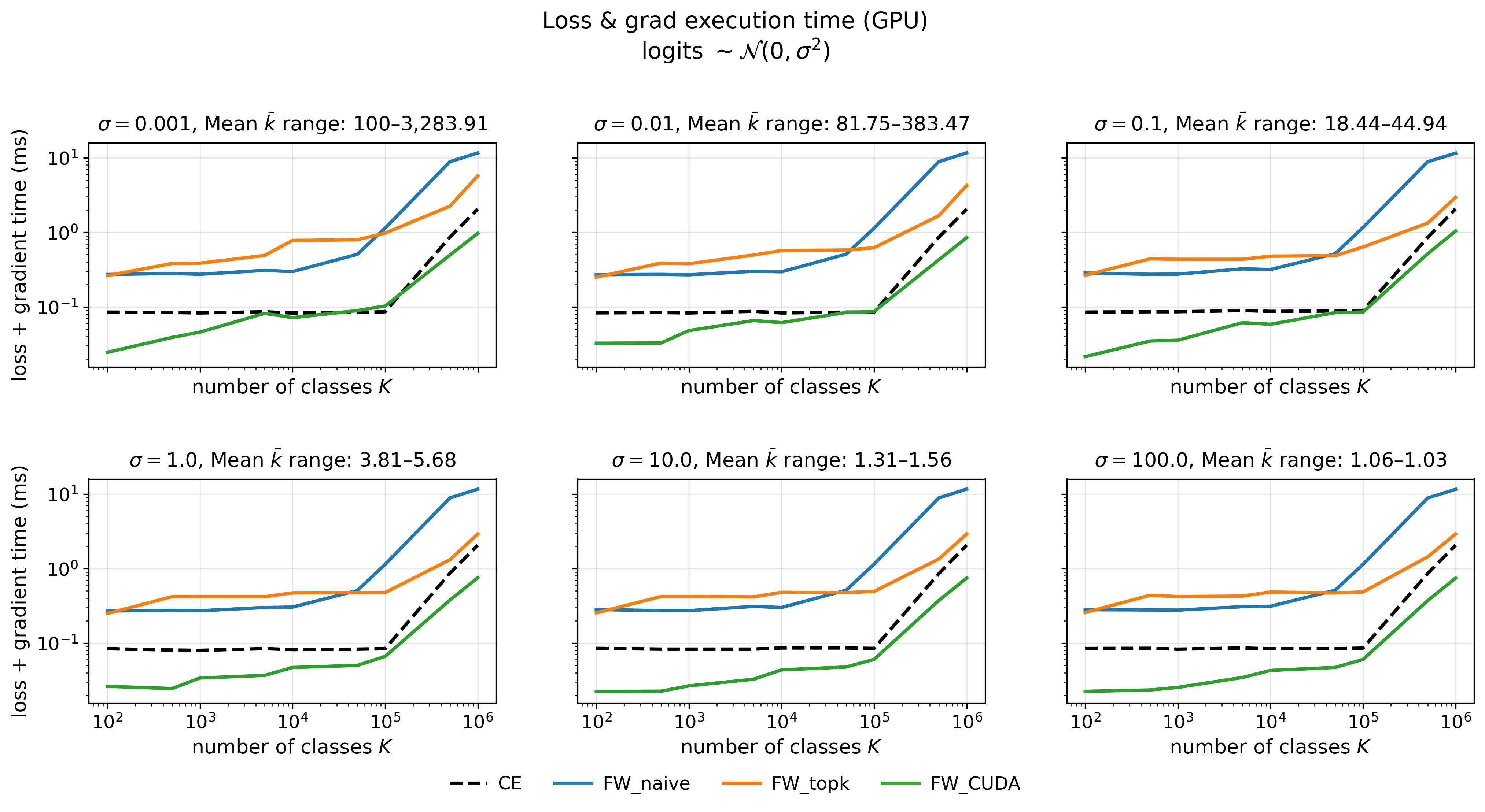}
  \caption{
  GPU runtime across different numbers of classes K and values of $\sigma$. Radix FW (FW\_CUDA), naive FW (FW\_naive), top-k FW (FW\_topk), and PyTorch cross-entropy (CE) are compared. Mean $\kappabar$ represents the average maximizing $\kappabar$ across $N=32$ samples for $K^2$ to $K^6$. Median execution time over 200 timed trials after 20 warm-up iterations.
  }
  \label{fig:GPU-time-comparison}
\end{figure*}
\begin{figure*}[t]
  \centering
\includegraphics[width=\textwidth]{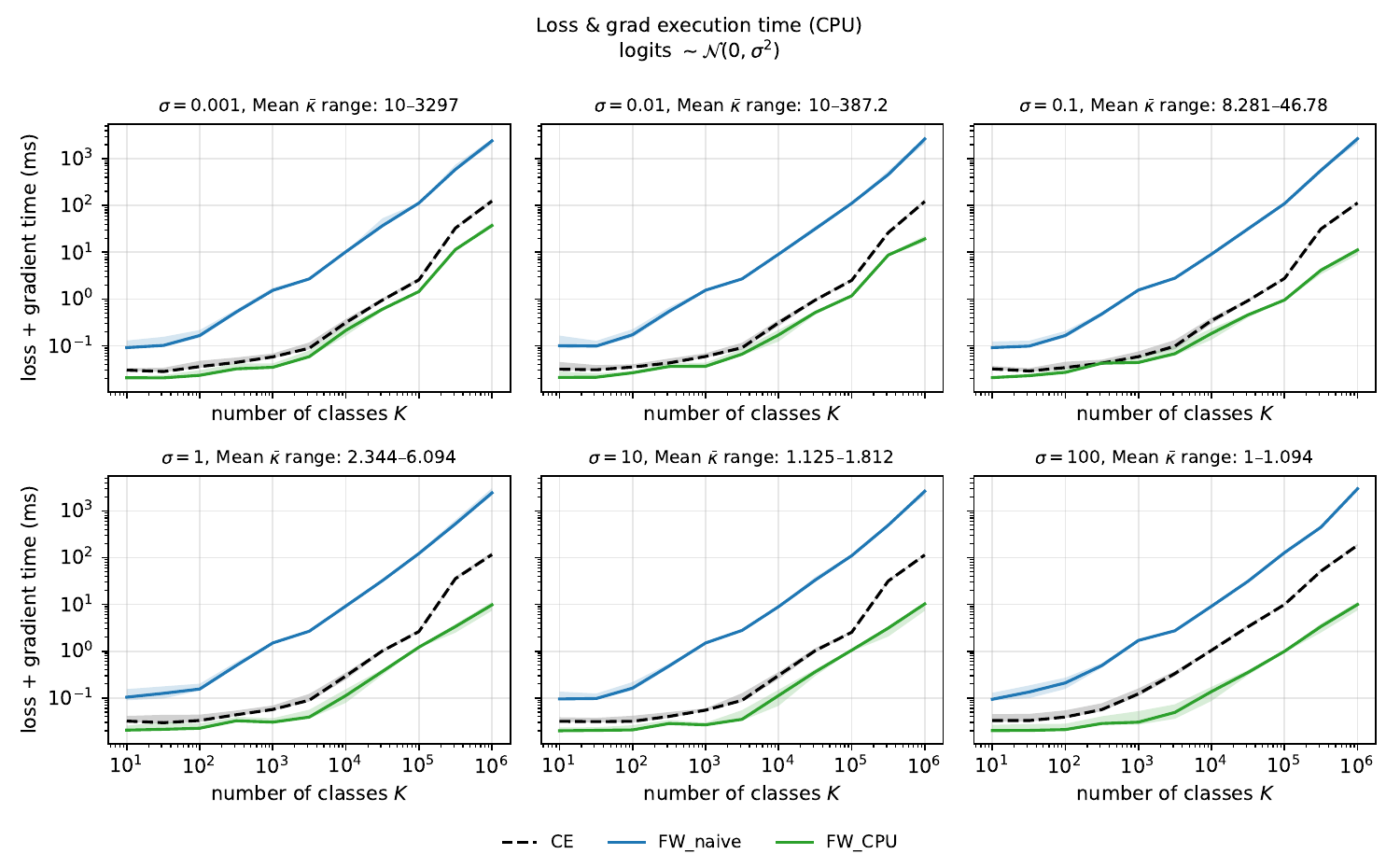}
  \caption{
  CPU loss and gradient computation time across different class counts K and sparsity regimes
  (batch size 32, two threads). We compare our single-pass threshold search (FW\_CPU) against PyTorch
  cross-entropy (CE) and the naive implementation (FW\_naive). Lines are medians and bands are the
  20--80 quantiles.
  }
  \label{fig:CPU-time-comparison}
\end{figure*}

\section{A Fast Algorithm for the Familywise loss}

The goal of this section is to establish theory for computing the familywise loss and gradient in \(O(K)\) time rather than \(O(K \log K)\). Concrete implementations will be
discussed in  \Cref{sec:implementations} below.
For reader's convenience, we restate the definition of the familywise loss which was introduced in Example~3 of \cite{duchi2018multiclass}:
\begin{definition}
  \label{definition:familywise-loss}
  Let \(\bfz \in \mathbb{R}^{K}\) and \(y \in \{1,\dots, K\}\). The \emph{familywise loss} is defined by
  \begin{equation}
    \textstyle
\calL(\bfz, y)
= 1 - z_{y} +
\max_{\kappa \in \{1,\dots, K\}} \left\{
  -\frac{1}{\kappa} +
\frac{1}{\kappa} \sum_{j=1}^{\kappa} z_{(j)}
\right\}
\label{equation:definition-of-familywise-loss}
\end{equation}
  where
                  \(
                  z_{(1)} \ge z_{(2)}
                  \ge \cdots \ge z_{(K)}
\)
is the entries of
\(\bfz\) sorted in descending order.
Let \(\kappabar(\bfz) \in \{1,\dots, K\}\) be the index that attains the maximum in
\eqref{equation:definition-of-familywise-loss}.
Throughout, whenever \(\bfz\) is unambiguous, we simply write \(\kappabar\) instead of \(\kappabar(\bfz)\).
\end{definition}

\begin{remark}[Naive algorithm]
When \(K\) becomes large, the \(\calO(K \log K)\) sorting operation becomes a computational bottleneck
when calculating  \(
\calL(\bfz, y)
\) directly using
\Cref{equation:definition-of-familywise-loss}.
We call this the ``naive'' algorithm/implementation and include the PyTorch code in \Cref{section:appendix:naive-impl-of-fwloss} using the autograd-enabled \verb|torch.sort| operator.
\end{remark}
\begin{definition}[\(\kappa\)-\shancs]
\label{equation:shancs}
The expression
inside ``max'' in
\eqref{equation:definition-of-familywise-loss}, i.e.,
\(H_{\kappa} :=
  -\frac{1}{\kappa} +
\frac{1}{\kappa} \sum_{j=1}^{\kappa} z_{(j)}\),
is referred to as  the \(\kappa\)-th \emph{sorted harmonically-normalized cumulative sum} (abbrev.\ \(\kappa\)-\shancs).
\end{definition}
As alluded to earlier, it is possible to avoid fully sorting all the logits, which becomes expensive for larger \(K\)'s. The key insight is that,
given knowledge of the \(\bar{\kappa} := \mathrm{argmax}_{\kappa} H_{\kappa}\)
and the ``threshold'' value \(z_{(\bar{\kappa})}\), it suffices to retrieve the top-\(\bar{\kappa}\) logits that are \(\ge z_{(\bar{\kappa})}\) which is a \(\mathcal{O}(K)\) task:
\begin{remark}\label{remark:prove-for-the-gradient-formula}
  Suppose that \(\kappabar\) attains the maximum in
\eqref{equation:definition-of-familywise-loss}, then the familywise loss  reduces to
\begin{equation}
  \label{equation:oracle-kappa-bar}
\textstyle
  \calL(\bfz,y)
  = 1 - z_{y} +
  - \frac{1}{\bar\kappa}+
          \frac{1}{\bar\kappa}   \sum_{j =1}^{\bar\kappa} z_{(j)}.
\end{equation}
Taking the gradient with respect to \(\bfz\), we get the formula~\eqref{equation:gradient-of-the-familywise-loss}
presented earlier in the introduction:
\begin{align}
  \label{equation:gradient-of-the-familywise-loss-main}
\nabla_{\bfz} \calL(\bfz, y)
  &= -\mbox{one-hot@\(y\)}
  \\ &\quad \,\,+ \,\,
  (1/ \bar{\kappa}) \cdot \mbox{many-hot@\(\{(1),\dots, (\bar{\kappa})\}\)} \notag
\end{align}
  where
  \(\mbox{one-hot@\(y\)}\) (resp.\
\(\mbox{many-hot@\(\{(1),\dots, (\bar{\kappa})\}\)}\))
denotes the \(K\)-dimensional vector with ``\(1\)'' at index \(y\) (resp.\
\(j\) for each
\(j \in
\{(1),\dots, (\bar{\kappa})\}
\)).
\end{remark}

The theory has three parts: an identity that removes the sorted order from the loss
(\Cref{theorem:threshold-identity}), its consequence that the loss and a subgradient cost
\(\calO(K)\) (\Cref{theorem:linear-time-computation-of-the-gradient}), and a bitonicity property
(\Cref{proposition:bitonicity-property-of-normalized-cumulative-sums}) on which our GPU kernel
relies.
 
\subsection{The familywise loss is a simplex-projection threshold}
\label{section:threshold}
For \(\bfz \in \bbR^{K}\) define the scalar function
\begin{equation}
  \label{equation:definition-of-g}
  \textstyle
  g_{\bfz}(t) := \sum_{j=1}^{K} \max(z_{j} - t,\, 0), \qquad t \in \bbR .
\end{equation}
The equation \(g_{\bfz}(t) = 1\) has exactly one solution, which we denote \(\tau^{*}(\bfz)\)
(\Cref{lemma:root}).
It is the threshold of the Euclidean projection of \(\bfz\) onto the probability simplex, whose
\(j\)-th coordinate is \(\max(z_{j} - \tau^{*}, 0)\)
\citep{held1974validation,michelot1986finite,duchi2008efficient,condat2016fast}; the same threshold
defines sparsemax \citep{martins2016softmax}.
 
\begin{theorem}[Threshold identity]
  \label{theorem:threshold-identity}
  Let \(\bfz \in \bbR^{K}\), let \(H_{\kappa}\) be the \(\kappa\)-\shancs
  (\Cref{equation:shancs}) and let \(\tau^{*} = \tau^{*}(\bfz)\). Then
  \begin{equation}
    \label{equation:threshold-identity}
    \max_{\kappa \in \{1,\dots,K\}} H_{\kappa} = \tau^{*},
    \quad\mbox{hence}\quad
    \calL(\bfz, y) = 1 - z_{y} + \tau^{*}.
  \end{equation}
  Moreover, \(\kappa\) attains the maximum if and only if
  \(|\{j : z_{j} > \tau^{*}\}| \le \kappa \le |\{j : z_{j} \ge \tau^{*}\}|\).
  In particular \(\kappabar := |\{j : z_{j} > \tau^{*}\}| \ge 1\) attains it,
  the top-\(\kappabar\) logits are exactly the logits that exceed \(\tau^{*}\), and
  \begin{equation}
    \label{equation:gradient-via-threshold}
    \tfrac{1}{\kappabar}\, \mathbf{1}[\bfz > \tau^{*}] \;-\; \mbox{one-hot@\(y\)}
  \end{equation}
  is a subgradient of \(\calL(\cdot, y)\) at \(\bfz\), where \(\mathbf{1}[\bfz > \tau^{*}]\) is the
  \(K\)-dimensional vector with ``\(1\)'' at every index \(j\) with \(z_{j} > \tau^{*}\).
  This is the vector in \eqref{equation:gradient-of-the-familywise-loss-main}.
\end{theorem}
\emph{Proof:} \Cref{section:appendix:threshold}.
\emph{Idea:} for every nonempty set \(S\) of classes,
\(\sum_{j \in S} (z_{j} - \tau^{*}) \le g_{\bfz}(\tau^{*}) = 1\), i.e.,
\(\frac{1}{|S|}(\sum_{j \in S} z_{j} - 1) \le \tau^{*}\), with equality for
\(S = \{j : z_{j} > \tau^{*}\}\).
The same argument shows that the adversarial zero-one loss of \cite{fathony2016adversarial}, a
maximum over all subsets \(S\), coincides with the FW loss (\eqref{equation:all-three-equal} in the
appendix).
 
\Cref{theorem:threshold-identity} changes the computational question from
``which \(\kappa\) is best?'', which refers to the sorted order, to
``where does the monotone function \(g_{\bfz}\) cross \(1\)?'', which does not.
 
\subsection{Linear-time computation}
\label{section:linear-time}
\begin{theorem}\label{theorem:linear-time-computation-of-the-gradient}
  Let \(\bfz \in \bbR^{K}\) and \(y \in \{1,\dots, K\}\) be arbitrary.
  The familywise loss \(\calL(\bfz, y)\) and a subgradient of \(\calL(\cdot, y)\) at \(\bfz\) can be
  computed in \(\calO(K)\) time in the worst case.
\end{theorem}
\emph{Proof:} \Cref{lemma:median-pivot} in \Cref{section:appendix:threshold}.
\emph{Idea:} by \Cref{theorem:threshold-identity} it suffices to find \(\tau^{*}\).
Since \(g_{\bfz}\) is non-increasing, \(g_{\bfz}(p) \ge 1\) if and only if \(p \le \tau^{*}\), so one
pass over the logits tells on which side of \(\tau^{*}\) any pivot \(p\) lies.
Taking \(p\) to be the median of the still undecided logits \citep{blum1973time}, each test settles
half of them: if \(g_{\bfz}(p) < 1\), those at or above \(p\) are known to exceed \(\tau^{*}\);
otherwise those at or below \(p\) are discarded.
The total work is \(\calO(K + K/2 + K/4 + \cdots) = \calO(K)\), and no sorted order is used.
 
Linear-time computation of the threshold \(\tau^{*}\) is classical in the optimization literature
\citep{brucker1984algorithm,kiwiel2008breakpoint,duchi2008efficient,condat2016fast}; what is new
here is \Cref{theorem:threshold-identity}, which makes these methods applicable to the familywise
loss. The median-based search has the right worst-case bound but a large constant, so our CPU
implementation (\Cref{sec:cpu-implementation}) uses a faster iteration and keeps the median-based
search only as a safeguard.
 
\subsection{Bitonicity}\label{section:bitonicity}
The second structural property concerns the sequence \(H_{1}, \dots, H_{K}\) itself: it first
increases and then decreases, so \(\kappabar\) is an index at which it turns.
 
\begin{proposition}[Bitonicity of \shancs]
  \label{proposition:bitonicity-property-of-normalized-cumulative-sums}
  Let \(z_{1},\dots, z_{K} \in \bbR\) be arbitrary numbers
  and
  \(z_{(1)} \ge \cdots \ge z_{(K)}\) be these numbers sorted in descending order.
Define
\(  H_{\kappa} :=
-\frac{1}{\kappa}
+\frac{1}{\kappa}
\left(
\sum_{j=1}^{\kappa} z_{(j)}
\right)
\).
Then \(H_{1},H_{2}, \dots\) is \emph{bitonic}, i.e., there exists \(\bar{\kappa} \in \{1,\dots, K\}\) such that
\[  H_{1} \le \dots \le H_{\bar{\kappa} - 1} \le H_{\bar{\kappa}}
  \ge
  H_{\bar{\kappa}+1} \ge \dots \ge H_{K}.\]
\end{proposition}
\emph{Proof:} \Cref{sec:proof-of-bitonicity}.
Our GPU kernel (\Cref{sec:cuda-kernel}) uses bitonicity as its stopping rule, in the form
\(H_{\kappa+1} > H_{\kappa}\) if and only if \(z_{(\kappa+1)} > H_{\kappa}\)
(\Cref{prop:stopping_condition}).
Bitonicity also gives a second proof of
\Cref{theorem:linear-time-computation-of-the-gradient}, by binary search over \(\kappa\)
(\Cref{section:appendix:second-proof}); the two proofs test the same condition
(\Cref{remark:same-test}).
 
\paragraph{Sparsity of the gradient.}
By \Cref{theorem:threshold-identity} the FW gradient has \(\kappabar + 1\) non-zero entries, whereas
the CE gradient is dense.
For Gaussian logits with standard deviation \(\sigma/K\), a heuristic calculation gives
\(\kappabar/K \approx (1/\sigma)\sqrt{2\log\sigma}\) for large \(K\) and \(\sigma\)
(\Cref{section:approximating-the-sparsity}), so the gradient becomes sparser as the logit scale
grows. None of our algorithms uses this estimate; it explains the sparsity that they exploit.
 
\section{Efficient implementations of familywise loss}
\label{sec:implementations}
By \Cref{theorem:threshold-identity}, the forward and the backward pass both reduce to two numbers
per sample: the threshold \(\tau^{*}\) and the count \(\kappabar\).
Our CPU implementation finds them in a single pass over the unsorted logits, without sorting,
heaps or top-\(\kappa\) selection (\Cref{sec:cpu-implementation}).
Our GPU implementation finds them by radix partitioning (\Cref{sec:cuda-kernel}).
 
\subsection{For CPU: single-pass threshold search}
\label{sec:cpu-implementation}
 
\paragraph{Algorithm.}
\Cref{alg:threshold-search} reads the logits once, in chunks of \(C = 2048\) that fit in the
first-level cache, while maintaining a lower bound \(t \le \tau^{*}\) and a buffer \(B\) that
contains every index seen so far with \(z_{j} > t\).
For each chunk it raises \(t\) to the chunk maximum minus one and appends the chunk entries above
\(t\) to \(B\).
Whenever \(B\) has grown by a factor of four, a few steps of the classical iteration
\citep{held1974validation,michelot1986finite}
\begin{equation}
  \label{equation:newton-step}
  \begin{aligned}
  t &\leftarrow t + \tfrac{1}{|B|}\bigl(\textstyle\sum_{j \in B} (z_{j} - t) - 1\bigr),\\
  B &\leftarrow \{ j \in B : z_{j} > t \}
  \end{aligned}
\end{equation}
raise \(t\) and shrink \(B\).
After the last chunk the iteration is run until \(B\) stops changing.
As a worst-case safeguard, if this final loop has visited \(4K\) entries without stopping, the
median-based search of \Cref{theorem:linear-time-computation-of-the-gradient} finishes on \(B\).
 
\begin{algorithm}[tb]
  \caption{FW loss and gradient by single-pass threshold search (one sample)}
  \label{alg:threshold-search}
  \begin{algorithmic}[1]
    \Require Logits $z[1 \ldots K]$, target $y$, chunk size $C$
    \State $t \leftarrow -\infty$;\quad $B \leftarrow \emptyset$;\quad $m \leftarrow 256$
    \For{each chunk $I$ of $C$ consecutive indices}
      \State $t \leftarrow \max\{t,\ \max_{j \in I} z_{j} - 1\}$
      \State $B \leftarrow B \cup \{j \in I : z_{j} > t\}$
      \If{$|B| \ge m$}
        \State $(t, B) \leftarrow \textsc{Tighten}(t, B, 4)$
        \State $m \leftarrow 4|B| + 256$
      \EndIf
    \EndFor
    \State $(t, B) \leftarrow \textsc{Tighten}(t, B, \infty)$
    \Statex \hspace{\algorithmicindent}(after $4K$ visited entries: median search on $B$)
    \State $\tau^{*} \leftarrow t$;\quad $\kappabar \leftarrow |B|$
    \State \textbf{loss:} $1 - z_{y} + \tau^{*}$
    \State \textbf{gradient:} $g \leftarrow 0$; $g_{j} \leftarrow 1/\kappabar$ for $j \in B$; $g_{y} \leftarrow g_{y} - 1$
    \Statex
    \Procedure{Tighten}{$t, B, r$}
      \State $B \leftarrow \{j \in B : z_{j} > t\}$
      \For{at most $r$ rounds}
        \State $t' \leftarrow t + \bigl(\sum_{j \in B} (z_{j} - t) - 1\bigr)/|B|$
        \State $B' \leftarrow \{j \in B : z_{j} > t'\}$
        \If{$B' = B$} \Return $(t', B)$ \EndIf
        \State $t \leftarrow t'$;\quad $B \leftarrow B'$
      \EndFor
      \State \Return $(t, B)$
    \EndProcedure
  \end{algorithmic}
\end{algorithm}
 
\begin{theorem}
  \label{theorem:cpu-algorithm}
  For every \(\bfz \in \bbR^{K}\) and \(y\), \Cref{alg:threshold-search} returns
  \(\tau^{*}(\bfz)\), \(\kappabar = |\{j : z_{j} > \tau^{*}\}|\), the loss
  \(\calL(\bfz, y)\) and the subgradient \eqref{equation:gradient-via-threshold}, in
  \(\calO(K)\) time in the worst case.
\end{theorem}
\emph{Proof:} \Cref{section:appendix:cpu-proof}.
\emph{Idea:} three consequences of \Cref{theorem:threshold-identity} are used.
\emph{(i)} \(g_{\bfz}(t) \ge 1\) if and only if \(t \le \tau^{*}\) (\Cref{lemma:root}).
\emph{(ii)} Every nonempty set \(S\) gives a lower bound,
\(\frac{1}{|S|}(\sum_{j \in S} z_{j} - 1) \le \tau^{*}\) (\Cref{lemma:subset-lower-bound}).
\emph{(iii)} If \(t \le \tau^{*}\), deleting every \(z_{j} \le t\) changes neither \(\tau^{*}\) nor
\(\{j : z_{j} > \tau^{*}\}\) (\Cref{lemma:restriction}).
Every value assigned to \(t\) has the form in (ii), with \(S\) a single index in the chunk step and
\(S = B\) in \eqref{equation:newton-step}, so \(t \le \tau^{*}\) throughout; an index leaves \(B\)
only when \(z_{j} \le t\), so by (iii) no logit above \(\tau^{*}\) is ever lost; and by (i) a step
of \eqref{equation:newton-step} never decreases \(t\) and can leave \(B\) unchanged only at
\(t = \tau^{*}\) (\Cref{lemma:newton}).
The buffer schedule makes the chunk phase linear, and the safeguard bounds the final loop.
No estimate of \(\kappabar\) is needed, and nothing is sorted.
 
\paragraph{Why it is fast.}
\emph{(i) The forward pass is comparisons}: a maximum and a ``keep the entries above \(t\)'' per
chunk, vectorized with AVX-512 or AVX2, whereas cross entropy evaluates an exponential for every
logit.
\emph{(ii) The backward pass is sparse}: the gradient \eqref{equation:gradient-via-threshold} has
\(\kappabar + 1\) non-zero entries whose indices the forward pass already holds in \(B\), whereas
cross entropy writes a dense gradient.
\emph{(iii) Each logit is read from memory once.}
Implementation details, including the floating-point safeguards, are in
\Cref{section:appendix:cpu-details}; measurements are in \Cref{sec:experiments}.

\subsection{For GPU: Custom CUDA Implementation via Radix Partitioning}
\label{sec:cuda-kernel}
Our GPU implementation incorporates the stopping condition of FW loss derived in Section \ref{prop:stopping_condition} into the radix partitioning algorithm to identify the maximizing \(\kappabar\); Unlike standard radix selection, which receives rank (k) as input and finds the (k)-th largest value,  our method uses the stopping condition to determine the number of logits required to maximize \(\kappa\)-\shancs. The algorithm partitions the logits into groups based on the successive bits of their order-preserving representation, allowing us to progressively refine the candidates contributing to 
\(\kappa\)-\shancs without fully sorting the logits. To support different logit dimensions K, two CUDA designs were developed while following the same underlying algorithm in Algorithm \ref{alg:radix-partitioning}. The execution strategy and memory organization are adjusted according to \(K\), as detailed later in this section.

\paragraph{Three-Stage CUDA Pipeline.}The implementation consists of three stages: 
Initial Threshold ($H$) Computation, Candidate Compaction, Radix Partitioning. The resulting procedure and a running example are given in Algorithm \ref{alg:radix-partitioning} and \ref{fig:gpu-running-example}, respectively.

\begin{algorithm}[t]
\caption{FW Loss via Radix Partitioning}
\label{alg:radix-partitioning}
\begin{algorithmic}[1]

\Require Logits $z[1,\ldots,K]$, target $y$, radix bits $r$

\State $\texttt{shift} \leftarrow 32-r$, $C \leftarrow 0$, $S \leftarrow 0$
\State Find threshold $H$
\State Compact logits $z_i > H$ into candidate buffer

\While{$\texttt{shift} \ge 0$}
    \State Build bin statistics Count, Sum, Min, Max
    \For{each bin $b$ in descending order}
        \State $C' \leftarrow C + \mathrm{Count}[b]$
        \State $S' \leftarrow S + \mathrm{Sum}[b]$
        \State $H_{C'} \leftarrow (S'-1)/C'$

        \If{$\mathrm{Min}[b] > H_{C'}$}
            \State $C \leftarrow C'$; $S \leftarrow S'$; $H \leftarrow H_{C'}$
            \State \textbf{continue to next bin}
        \ElsIf{$\mathrm{Max}[b] > H_{C'}$}
            \State Refine bin $b$; \textbf{break}
        \Else
            \State \textbf{break}
        \EndIf
    \EndFor
    \State $\texttt{shift} \leftarrow \texttt{shift}-r$
\EndWhile
\State $\mathcal{L}_{\mathrm{FW}} \leftarrow 1-z_y+H$
\end{algorithmic}
\end{algorithm}

\paragraph{Stage 1: Initial Threshold (H) Computation.} The Proposition \ref{proposition:bitonicity-property-of-normalized-cumulative-sums} shows that $H_k$ is non-decreasing up to the maximizing $\kappabar$. Therefore, computing $H_1 = z_1 - 1$ provides a lower bound for all subsequent $H_2, H_3, ...,H_{\kappabar}$. We compute $H_1$ by finding $z_1$, the largest logit, using a $max$ operation. Also, as shown in Proposition \ref{prop:stopping_condition}, $H_{k+1} > H_k$ if and only if $z_{k+1} > H_k$. Thus,
logits satisfying $z_i <= H_1$ can be removed, since they cannot increase $H_k$ later in the sequence. For $K \le 10K $, we use $H_1$ as the threshold for candidate filtering. However, for larger $K$, $H_1$ may not be a sufficiently tight lower bound and can leave many candidate logits. Therefore, for $K > 10K$, we perform one radix pass over the logits and progressively update $H_c$ as bins are accepted. When the algorithm first enters the refine case, the current $H_c$ is used as a tighter threshold for candidate filtering. This further reduces the candidate set before the remaining radix passes.

\paragraph{Stage 2: Candidate Compaction.} For smaller cases, $K \le 10K$, this stage creates the candidate buffer in GPU shared memory, since it provide faster access than GPU global memory. For larger cases, $K > 10K$, the candidate buffer is maintained in GPU global memory to accommodate the larger set of candidate logits. The  choice of $10K$ is due to the limited shared-memory capacity available to each CUDA block. For targe K, the candidate buffer may not fit efficiently in GPU shared memory, so it is stored in GPU global memory instead.

\paragraph{Stage 3: Radix Partitioning.} Basically, the radix algorithm requires a key representation whose bit order is consistent with the numerical order. Therefore, each reduced candidate logit is mapped to an order-preserving key using the standard bitwise transformation of the IEEE-754 floating-point representation such that comparing the key values is equivalent to comparing the original logits. Our algorithm \ref{alg:radix-partitioning} starts scanning these order-preserving keys from the MSB (Most Significant Bit), considering the $r$ bits in each pass, where r denotes the number bits processed per pass. At each pass, the logits are grouped into $2^r$ bins using the current prefix and the next $r$ bits, while accumulating Count, Sum, Min, and Max for each bin. These bin statistics are evaluated against the FW Loss stopping condition \ref{prop:stopping_condition} to decide among the three cases :  accept the bin and move to the next bin, refine the bin using the next $r$ bits, or stop the search. This determines the maximizing ${\kappabar}$ without explicit sorting or rearranging the candidate logits.

\begin{figure*}[t]
  \centering
\includegraphics[width=0.32\textwidth]{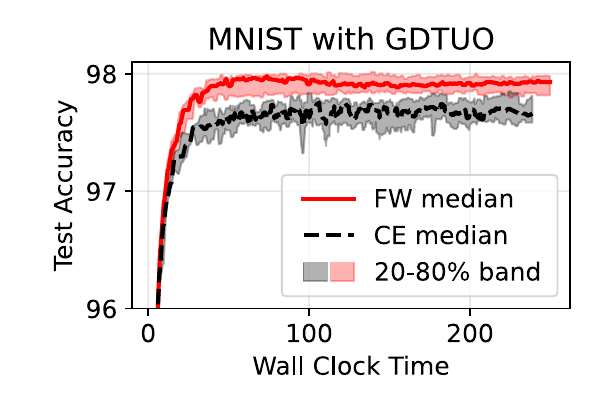}~
\includegraphics[width=0.32\textwidth]{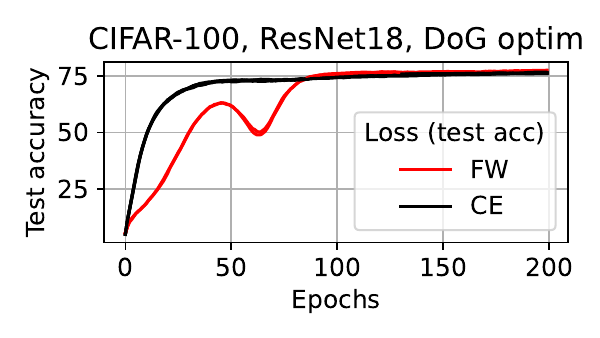}
~
\includegraphics[width=0.32\textwidth]{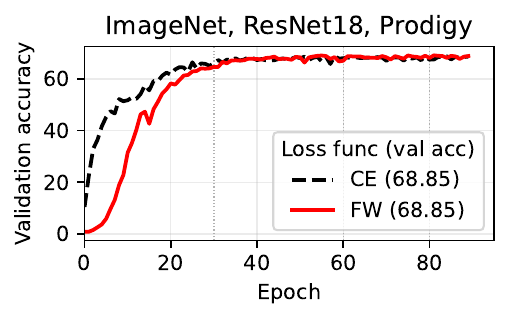}
\caption{
  (\emph{Left}) Test accuracy vs wall clock time on MNIST using the example code provided by GDTUO \citep{chandra2022gradient}.
  (\emph{Middle}) 
    Test accuracy versus epoch on training ResNet18 using the DoG optimizer \citep{ivgi2023dog}.
  (\emph{Right}) Validation accuracy vs epoch on ImageNet using Prodigy \cite{mishchenko2024prodigy}. Total wall clock times for both training runs are \(\sim\)48600 seconds.
}
\label{fig:mnist-gdtuo}
\end{figure*}

\paragraph{CUDA Design.} We use one CUDA block per input sample for candidate compaction when $K \le 10K$, with the number of threads chosen based on K. For larger cases, $K > 10K$, the compaction stage uses multiple CUDA blocks per input sample, with 256 threads per block, to parallelize the filtering operation. As described in Stage 2, the compacted candidates are written to the appropriate memory location based on K: shared memory for smaller K and global memory for larger K. The radix-partitioning stage then processes the remaining candidate logits using a single CUDA block per input sample. Within this block, threads cooperatively process the candidate logits, while per-warp radix histograms accumulate the Count, Sum, Min, and Max statistics. We use r=4 for $K \le 10K$ and r=8 for larger K. When a bin is selected for refinement, its radix bits are appended to the current prefix, and the next pass processes only the candidates whose keys match that updated prefix.

\paragraph{Complexity.} The initial-threshold computation and candidate compaction each require a single pass over the \(K\) logits, giving \(O(K)\) work. Each radix pass processes the candidate logits once and accumulates statistics for $2^r$ bins. Let M denote the number of candidate logits after compaction. One radix pass requires $O(M)$ work. Since $M \le K$, this is $O(K)$ in the worst case. For fixed $r$, the number of bins is constant, and because FP32 keys contain 32 bits, the number of radix passes is at most $32/r$. Therefore, the overall procedure has $O(K)$ worst-case time complexity.  

\section{Experiments} 
\label{sec:experiments}

We evaluate the proposed GPU and CPU implementations in terms of correctness and runtime performance across different logit dimensions K and input distributions. 

\paragraph{Benchmarking data.} We sample\footnote{We also tested logits distributed from a mixture of Gaussian and Laplacian. Figures are included in the Appendix\ref{section:appendix:distributions}. } logits independently from $\mathcal{N}(0,\sigma^2)$, with $\sigma$ ranging from $10^{-3}$ to $10^2$. We use a batch size of $N = 32$ and vary the number of classes from $10$ to $10^6$. For each sample, the target class is chosen uniformly at random from $\{0, ..., K-1\}$. 

\paragraph{Baselines.} For the GPU experiments, we compare the runtime of our radix FW implementation(\ref{alg:radix-partitioning}) against three baselines: the naive FW, the top-k FW, and PyTorch cross-entropy (CE). For the CPU runtime comparison, we use naive FW and cross-entropy (CE).

\paragraph{Correctness.} To verify  implementation, 
we use the naive FW implementation as the exact reference and compare the loss and gradients checked using relative tolerance of $10^{-5}$ and absolute tolerance of $10^{-6}$. 

\subsection{Runtime Benchmarking}
\paragraph{GPU.} Figure \ref{fig:GPU-time-comparison} shows the runtime of radix FW compared with naive FW, top-k, and PyTorch cross-entropy across different values of $K$ and $\sigma$. Radix FW achieves the lowest runtime across all tested configurations. For smaller values of K, roughly from $10^2$ to $10^3$, its runtime remains below 0.1 ms across all distributions and is consistently faster than CE, while naive FW and top-k FW are already noticeably more expensive. As K increases, the runtime advantage of radix FW over the other FW baselines becomes more pronounced. Naive FW becomes much more expensive beyond $K=10^5$, mainly due to sorting the full logit vector, while top-k FW scales better than naive FW but still remains slower than radix FW across the tested range.

Radix FW also remains faster than CE across all evaluated values of K. Up to approximately $K=10^5$, both radix FW and CE remain below 0.1 ms in most settings, with radix FW consistently lower. For larger K, the runtime of both methods increases, but radix FW maintains a clear advantage. At $K=10^6$, radix FW remains below 1 ms across the tested distributions, while CE is around 2 ms and the existing FW baselines are considerably slower. As K grows further, CE approaches the runtime of top-k FW, while radix FW continues to scale more gradually.

\paragraph{CPU.}
\Cref{fig:CPU-time-comparison} shows the time of one loss-and-gradient evaluation
(\texttt{loss(logits, target).backward()} with \texttt{reduction=`sum'}; FW and CE are timed
alternately under the same protocol).
Our implementation is faster than CE in all \(66\) configurations.
The ratio of the median times (CE over FW) is \(1.22\)--\(4.17\) for \(K \le 10^{3}\),
\(1.43\)--\(9.64\) for \(10^{3} < K \le 10^{5}\), and \(2.88\)--\(16.18\) for \(K > 10^{5}\).
At \(K = 10^{6}\), FW takes \(11.9\)--\(41.1\)\,ms and CE \(125\)--\(203\)\,ms.
The 20--80 quantile bands of the two methods are disjoint in \(53\) configurations; the \(13\) in
which they overlap all have \(K \le 10^{4}\).
Against naive FW, ours is faster in all \(66\) configurations, by a factor between \(4.3\) and
\(218.2\).
The advantage over CE has two sources: the forward pass consists of comparisons instead of
exponentials, and the backward pass writes only the \(\kappabar + 1\) non-zero entries of the
gradient.
\paragraph{Effect of $\sigma$.} Both CPU and GPU implementations show consistent runtime behavior across different values of \(\sigma\), indicating robustness to changes in the input distribution. We explain the details in the Appendix.~\ref{section:appendix:sigma-effect}

\subsection{Evaluation on real datasets}
Evaluation of the familywise loss was done in 
\cite{fathony2016adversarial} showed that the familywise loss is superior to other multiclass losses for the SVM.
Here we show that a familywise loss trained neural network
using
parameter-free learning methods
outperforms cross-entropy.
We refer the reader to appendix \ref{section:appendix:experiments} for  detailed results.

\section{Discussion}
\label{sec:discussion}

This paper continues the study of familywise loss initiated by \cite{duchi2018multiclass,fathony2016adversarial} as an effective alternative to cross-entropy in large-scale classification.
While earlier papers focused on statistical theory, our work focuses theory for efficient computation.
Below, we discuss future directions.

There is a revival in interest in sparsity for deep learning \citep{martins2016softmax,nikdan2023sparseprop}.
However, it is folklore that sparsity only pays off at extreme level of sparsity.
Here, we observed during training
that the
familywise loss produces gradients that naturally become increasingly more sparse. This is in contrast to  post-hoc gradient pruning methods
that modifies a (dense) gradient into a sparse ``pseudo''-gradient.
Our theory result in
\Cref{section:approximating-the-sparsity}
allows estimation of the sparsity of the gradient
and opens the door to
detect the sparsity threshold for switching to sparse matrix-vector multiply dynamically.

Sparsity may also be helpful
for
alleviating memory burden
large language models pretraining when the number of classes, i.e., the vocabulary size, is large \cite{wijmans2025cut}.
Moreover, there is increasing evidence that increasing the size of the vocabulary can be helpful.
See
\citet[Fig.~4]{geiping2023cramming}
which demonstrates that the mean GLUE score increases with vocabulary size in a BERT-variant
and more recently
\cite{tao2024scaling}
for larger language models.
Future work will explore the role of familywise loss in the vocabulary-size scaling regime.

\section*{Impact Statement}

This paper presents work whose goal is to advance the field of loss function theory. There are many potential societal consequences of our work, none
which we feel must be specifically highlighted here.

\subsection*{AI use statement}
In this work, we used generative AI tools to help formulate mathematical claims and to engineer the kernel and its tests, and to design
the CPU timing experiments on randomly generated logits and summarize their results.
We have not used generative AI tools to develop the familywise loss itself, which is from prior
work, or to run or analyze the training experiments on real datasets.
Translation, dataset cleaning, and qualitative data analysis are not applicable to this work.
Additionally, we used generative AI tools to create plotting and benchmarking code and figures, to
search for and summarize related literature, and to edit parts of the text, especially grammar.
AI-generated code was tested against an independent double-precision reference,
and the literature claims were
checked against the cited sources.
We take responsibility for the final content of this work, including text, claims, or artifacts
produced with the aid of generative AI.

\subsubsection*{Acknowledgements}
HB is supported by JST PRESTO JPMJPR24K6.
YW acknowledges support from NSF CISE CRII 2451714 and NSF CCF 2551504.
A part of the experiments of this research was conducted using Wisteria/Aquarius in the Information
Technology Center, the University of Tokyo.

\bibliography{references_YW,references_HB}

\clearpage
\appendix
\thispagestyle{empty}

\onecolumn
\aistatstitle{A Fast Algorithm for the Familywise Loss: \\
Supplementary Materials}

\section{Proof of results from Section 3}

\subsection{Proof of \texorpdfstring{\Cref{proposition:bitonicity-property-of-normalized-cumulative-sums}}{proposition}}

\label{sec:proof-of-bitonicity}

We restate the proposition for the reader's convenience.
\begin{proposition}[Bitonicity]\label{proposition-repeated:bitonicity-property-of-normalized-cumulative-sums}
  Let \(z_{1},\dots, z_{K} \in \bbR\) be arbitrary numbers
  and
  \(z_{(1)} \ge \cdots \ge z_{(K)}\) be these numbers sorted in descending order.
Define
\(  R_{\ell} :=
  z_{(1)} + \cdots + z_{(\ell)}
  =
\sum_{j=1}^{\ell} z_{(j)}
\)
and
\(  H_{\ell} :=
\frac{1}{\ell} \left(R_{\ell}- 1\right)
\).

Then \(H_{1},H_{2}, \dots\) is \emph{unimodal}, i.e., there exists \(m^{*} \in \{1,\dots, K\}\) such that
\[  H_{1} \le \dots \le H_{m^{*} - 1} \le H_{m^{*}}
  \ge
  H_{m^{*}+1} \ge \dots \ge H_{K}.\]
\end{proposition}

\begin{proof}
   To prove the proposition, we note that the result is equivalent to the following statement:
There exists at most one \(m \in \{1,\dots, K-1\}\) such that
\[
(H_{m+1}  - H_{m}) (H_{m} - H_{m_{1}}) < 0
\]
For simplicity, let's write \(s_{\ell} := z_{(\ell)}\).
Now, we calculate \(H_{m+1} - H_{m}\):
\begin{align}
  H_{m+1} - H_{m}
  &=
  \frac{R_{m+1} - 1}{m+1}
  -
  \frac{R_{m} - 1}{m}
    \\
  &=
  \frac{m(R_{m}+s_{m+1}) - m - (m+1)R_{m} + (m+1)}{m(m+1)}
  \\
  &=
  \frac{ m s_{m+1} - R_{m} + 1}{m(m+1)}.
\end{align}
Below, we focus on the above numerator.
The claim is that
\[
  \delta_{m} := m s_{m+1} - R_{m} + 1
\]
is non-increasing as \(m\) increases, and hence the sequences \(\delta_{1},\delta_{2},\dots\) changes sign only once.
To see this, note that
\begin{align*}
&\delta_{m+1} - \delta_{m}
  \\
  &=
 (m+1)s_{m+2}  - R_{m+1} +1  - (m s_{m+1}- R_{m} +1)
    \\
 &=
 (m+1)s_{m+2}  - s_{m+1} - m s_{m+1}
   \\
 &=
 (m+1)s_{m+2}   - (m+1) s_{m+1}
   \\
&=
 (m+1)(s_{m+2}   - s_{m+1}) \le 0
\end{align*}
Thus, \(\delta_{m}\) is non-increasing in \(m\).
\end{proof}

\begin{proposition}\label{prop:stopping_condition}
  Let \(z_{1},\dots, z_{K} \in \bbR\) be arbitrary numbers
  and
  \(z_{(1)} \ge \cdots \ge z_{(K)}\) be these numbers sorted in descending order.
Define
\(  S_{\kappa} :=
  z_{(1)} + \cdots + z_{(\kappa)}
  =
\sum_{j=1}^{\kappa} z_{(j)}
\)
and
\(  H_{\kappa} :=
\frac{S_{\kappa}-1}{\kappa}
\). 

Then, \(H_{k+1} > H_{k} \quad \Longleftrightarrow \quad Z_{k+1} > H_{k}.\)

\begin{proof}
Since \(S_{k+1} = S_k + z_{(k+1)}\),
we have \(H_{k+1} = \frac{S_k + z_{(k+1)} -1}{k+1}\).
Therefore,
\begin{align*}
    H_{k+1} > H_k &\Longleftrightarrow \frac{S_k+z_{(k+1)}-1}{k+1} > \frac{S_k-1}{k} \\
    &\Longleftrightarrow k\bigl(S_k+z_{(k+1)}-1\bigr)
    > (k+1)(S_k-1) \\
    &\Longleftrightarrow kz_{(k+1)} > S_k-1 \\
    &\Longleftrightarrow z_{(k+1)} > \frac{S_k-1}{k} = H_k.
\end{align*}
\end{proof}
\end{proposition}

\subsection{A second proof of \Cref{theorem:linear-time-computation-of-the-gradient}: binary search over \(\kappa\)}
\label{section:appendix:second-proof}
This proof uses bitonicity and does not use the threshold \(\tau^{*}\) of
\Cref{theorem:threshold-identity}.
We keep the notation \(S_{\kappa} = \sum_{j=1}^{\kappa} z_{(j)}\) and
\(H_{\kappa} = (S_{\kappa} - 1)/\kappa\) of \Cref{prop:stopping_condition}, and for a nonempty set
\(S\) of indices we write \(H(S) := \frac{1}{|S|}\bigl(\sum_{j \in S} z_{j} - 1\bigr)\).
A \emph{maximizer} is an index \(\kappa\) with \(H_{\kappa} = \max_{\kappa'} H_{\kappa'}\).
 
\begin{lemma}[Peak test]
  \label{lemma:peak-test}
  Let \(m \in \{1, \dots, K-1\}\).
  \begin{enumerate}
    \item[(a)] If \(z_{(m+1)} > H_{m}\), then every maximizer is \(\ge m+1\).
    \item[(b)] If \(z_{(m+1)} \le H_{m}\), then \(H_{\kappa+1} \le H_{\kappa}\) for every
      \(\kappa \in \{m, \dots, K-1\}\); in particular \(H_{m} \ge H_{\kappa}\) for every \(\kappa \ge m\).
  \end{enumerate}
\end{lemma}
\begin{proof}
For every \(\kappa < K\) we have \(S_{\kappa+1} - 1 = \kappa H_{\kappa} + z_{(\kappa+1)}\), hence
\begin{equation}
  \label{equation:convex-combination}
  H_{\kappa+1} = \frac{\kappa}{\kappa+1}\, H_{\kappa} + \frac{1}{\kappa+1}\, z_{(\kappa+1)} ,
\end{equation}
a convex combination of \(H_{\kappa}\) and \(z_{(\kappa+1)}\). In particular
\(H_{\kappa+1} \le H_{\kappa}\) if and only if \(z_{(\kappa+1)} \le H_{\kappa}\)
(this is \Cref{prop:stopping_condition}).
 
(b) We show by induction that \(z_{(\kappa+1)} \le H_{\kappa}\) for every
\(\kappa \in \{m, \dots, K-1\}\), which gives the claim.
For \(\kappa = m\) this is the hypothesis.
If \(z_{(\kappa+1)} \le H_{\kappa}\) and \(\kappa + 1 < K\), then by
\eqref{equation:convex-combination} \(H_{\kappa+1} \ge z_{(\kappa+1)} \ge z_{(\kappa+2)}\).
 
(a) By \eqref{equation:convex-combination}, \(H_{m+1} > H_{m}\), so \(m\) is not a maximizer.
Suppose some \(\kappa < m\) were a maximizer. Then \(H_{\kappa+1} \le H_{\kappa}\), i.e.,
\(z_{(\kappa+1)} \le H_{\kappa}\), and part (b) applied to \(\kappa\) gives
\(H_{m+1} \le H_{m}\), a contradiction.
\end{proof}
\Cref{lemma:peak-test} is the form of
\Cref{proposition:bitonicity-property-of-normalized-cumulative-sums} that the search needs: one
comparison at \(m\) tells on which side of \(m\) a maximizer lies, including when there are ties.
 
\begin{proof}[Second proof of \Cref{theorem:linear-time-computation-of-the-gradient}]
\emph{Algorithm.}
We maintain integers \(1 \le \ell \le u \le K\), a number \(s\) and a multiset \(U\) of logits such that
\begin{enumerate}
  \item[(I1)] some maximizer lies in \(\{\ell, \dots, u\}\);
  \item[(I2)] \(U = \{z_{(\ell)}, \dots, z_{(u)}\}\) as a multiset, and \(s = S_{\ell-1}\).
\end{enumerate}
Initially \(\ell = 1\), \(u = K\), \(s = 0\) and \(U\) contains all logits.
While \(\ell < u\), let \(n := |U| = u - \ell + 1 \ge 2\), \(q := \lfloor n/2 \rfloor\) and
\(m := \ell - 1 + q\), so that \(\ell \le m < u\).
Find the \(q\)-th largest element of \(U\) in \(\calO(n)\) time \citep{blum1973time} and split
\(U\) into the multiset \(U^{+}\) of its \(q\) largest elements and the multiset \(U^{-}\) of the
other \(n - q\) elements (ties are split arbitrarily).
By (I2), \(U^{+} = \{z_{(\ell)}, \dots, z_{(m)}\}\) and \(U^{-} = \{z_{(m+1)}, \dots, z_{(u)}\}\),
so one pass over \(U\) gives
\[
  S_{m} = s + \textstyle\sum_{x \in U^{+}} x, \qquad H_{m} = (S_{m} - 1)/m, \qquad z_{(m+1)} = \max U^{-} .
\]
\begin{itemize}
  \item If \(z_{(m+1)} > H_{m}\): by \Cref{lemma:peak-test}(a) every maximizer is \(\ge m+1\), so by
    (I1) one lies in \(\{m+1, \dots, u\}\). Set \(\ell \leftarrow m+1\), \(s \leftarrow S_{m}\),
    \(U \leftarrow U^{-}\).
  \item Otherwise: by \Cref{lemma:peak-test}(b), \(H_{m} \ge H_{\kappa}\) for all \(\kappa \ge m\).
    If the maximizer of (I1) is larger than \(m\), then \(m\) is a maximizer as well; in both cases
    a maximizer lies in \(\{\ell, \dots, m\}\). Set \(u \leftarrow m\), \(U \leftarrow U^{+}\).
\end{itemize}
Both updates preserve (I1) and (I2).
 
\emph{Running time.}
A round costs \(\calO(n)\) and replaces \(n\) by \(n - q = \lceil n/2 \rceil\) or by
\(q \le \lceil n/2 \rceil\). Hence the size of \(U\) in round \(t\) is at most
\(\lceil K/2^{t} \rceil\), there are at most \(\lceil \log_{2} K \rceil\) rounds, and the total
cost is \(\calO\bigl(\sum_{t \ge 0} \lceil K/2^{t} \rceil\bigr) = \calO(2K + \log_{2} K) = \calO(K)\).
No assumption on \(K\) is needed.
 
\emph{Output.}
When the loop ends, \(\kappabar := \ell = u\) is a maximizer by (I1), \(U = \{v\}\) with
\(v = z_{(\kappabar)}\) by (I2), and \(\max_{\kappa} H_{\kappa} = (s + v - 1)/\kappabar\), which gives
the loss.
For the subgradient, one more pass over \(\bfz\) collects \(A := \{j : z_{j} > v\}\), which has fewer
than \(\kappabar\) elements, and \(\kappabar - |A|\) further indices \(j\) with \(z_{j} = v\); call the
resulting set \(T\). Then \(T\) is a set of \(\kappabar\) largest entries, so \(H(T) = H_{\kappabar}\).
Since the sum of the \(\kappa\) largest entries is the maximum of \(\sum_{j \in S} z_{j}\) over sets
\(S\) with \(|S| = \kappa\), we have \(\calL(\bfz, y) = 1 - z_{y} + \max_{S \ne \emptyset} H(S)\), a
maximum of affine functions of \(\bfz\), and \(T\) attains it. The gradient of an affine function
attaining the maximum is a subgradient, which here is
\(\frac{1}{\kappabar}\mathbf{1}_{T} - \mbox{one-hot@\(y\)}\), the vector in
\eqref{equation:gradient-of-the-familywise-loss-main}.
\end{proof}
 
\begin{remark}[The two proofs test the same condition]
  \label{remark:same-test}
  With \(g_{\bfz}\) as in \eqref{equation:definition-of-g}, only the \(m\) largest logits contribute
  to \(g_{\bfz}(z_{(m+1)})\), so
  \[
    g_{\bfz}\bigl(z_{(m+1)}\bigr) = \sum_{j=1}^{m} \bigl(z_{(j)} - z_{(m+1)}\bigr)
    = S_{m} - m\, z_{(m+1)} = 1 + m\,\bigl(H_{m} - z_{(m+1)}\bigr).
  \]
  Hence \(z_{(m+1)} > H_{m}\) if and only if \(g_{\bfz}(z_{(m+1)}) < 1\), i.e., if and only if
  \(z_{(m+1)} > \tau^{*}\) (\Cref{lemma:root}).
  The comparison made by the binary search at rank \(m\) is therefore the test made by the
  median-based search of \Cref{lemma:median-pivot} at the pivot \(z_{(m+1)}\).
  The two proofs differ in how they justify discarding half of the logits: by the bitonicity of
  \(\kappa \mapsto H_{\kappa}\) in one case, and by the monotonicity of \(t \mapsto g_{\bfz}(t)\) in
  the other.
\end{remark}

\section{Proofs for the threshold identity and the CPU algorithm}
\label{section:appendix:threshold}
Throughout, \(\bfz \in \bbR^{K}\) with \(K \ge 1\), \((a)_{+} := \max(a, 0)\),
\(g(t) := g_{\bfz}(t) = \sum_{j=1}^{K} (z_{j} - t)_{+}\), and \(z_{\max} := z_{(1)}\).
For a nonempty \(S \subseteq \{1, \dots, K\}\) put
\(H(S) := \frac{1}{|S|}\bigl(\sum_{j \in S} z_{j} - 1\bigr)\), so that \(H_{\kappa} = H(S)\) for
any set \(S\) of indices of \(\kappa\) largest entries.
 
\begin{lemma}[The root \(\tau^{*}\)]
  \label{lemma:root}
  The function \(g\) is continuous and non-increasing on \(\bbR\) and strictly decreasing on
  \((-\infty, z_{\max}]\). There is a unique \(\tau^{*}\) with \(g(\tau^{*}) = 1\), and
  \(z_{\max} - 1 \le \tau^{*} < z_{\max}\).
  Consequently, \(g(t) \ge 1\) if and only if \(t \le \tau^{*}\).
\end{lemma}
\begin{proof}
  Each \(t \mapsto (z_{j} - t)_{+}\) is continuous and non-increasing, hence so is \(g\).
  Let \(j_{0}\) be an index with \(z_{j_{0}} = z_{\max}\).
  For \(t_{1} < t_{2} \le z_{\max}\), the \(j_{0}\)-th term decreases by exactly \(t_{2} - t_{1}\)
  and no other term increases, so \(g(t_{1}) - g(t_{2}) \ge t_{2} - t_{1} > 0\).
  Next, \(g(z_{\max}) = 0\) and \(g(z_{\max} - 1) \ge (z_{j_{0}} - z_{\max} + 1)_{+} = 1\).
  By the intermediate value theorem there is \(\tau^{*} \in [z_{\max} - 1, z_{\max})\) with
  \(g(\tau^{*}) = 1\).
  It is unique because \(g\) is strictly decreasing on \((-\infty, z_{\max}]\) and identically
  \(0\) on \([z_{\max}, \infty)\).
  For the last claim: if \(t \le \tau^{*}\) then \(g(t) \ge g(\tau^{*}) = 1\) by monotonicity; if
  \(t > \tau^{*}\) then either \(t \le z_{\max}\) and \(g(t) < g(\tau^{*}) = 1\) by strict
  monotonicity, or \(t > z_{\max}\) and \(g(t) = 0 < 1\).
\end{proof}
 
\begin{lemma}[Every subset is a lower bound]
  \label{lemma:subset-lower-bound}
  For every nonempty \(S \subseteq \{1, \dots, K\}\), \(H(S) \le \tau^{*}\).
\end{lemma}
\begin{proof}
  \(\sum_{j \in S} (z_{j} - \tau^{*}) \le \sum_{j \in S} (z_{j} - \tau^{*})_{+}
  \le \sum_{j=1}^{K} (z_{j} - \tau^{*})_{+} = g(\tau^{*}) = 1\).
  Rearranging, \(\sum_{j \in S} z_{j} - 1 \le |S|\, \tau^{*}\), i.e., \(H(S) \le \tau^{*}\).
\end{proof}
 
\begin{proof}[Proof of \Cref{theorem:threshold-identity}]
  Let \(S^{*} := \{j : z_{j} > \tau^{*}\}\) and \(c := |S^{*}|\).
  By \Cref{lemma:root}, \(\tau^{*} < z_{\max}\), so \(c \ge 1\).
 
  \emph{Value.}
  Only indices in \(S^{*}\) contribute to \(g(\tau^{*})\), so
  \(\sum_{j \in S^{*}} (z_{j} - \tau^{*}) = g(\tau^{*}) = 1\), i.e., \(H(S^{*}) = \tau^{*}\).
  Every entry indexed by \(S^{*}\) is larger than every entry not indexed by \(S^{*}\), so \(S^{*}\)
  is the set of indices of the \(c\) largest entries and \(H_{c} = H(S^{*}) = \tau^{*}\).
  Together with \Cref{lemma:subset-lower-bound},
  \begin{equation}
    \label{equation:all-three-equal}
    \max_{\kappa \in \{1,\dots,K\}} H_{\kappa}
    \;=\; \max_{\emptyset \ne S \subseteq \{1,\dots,K\}} H(S)
    \;=\; \tau^{*},
  \end{equation}
  and the formula for \(\calL\) follows from \eqref{equation:definition-of-familywise-loss}.
 
  \emph{Maximizers.}
  Let \(c_{\ge} := |\{j : z_{j} \ge \tau^{*}\}|\) and let \(T\) be a set of indices of \(\kappa\)
  largest entries, so that \(\kappa (H_{\kappa} - \tau^{*}) = \sum_{j \in T} (z_{j} - \tau^{*}) - 1\).
  If \(\kappa < c\), then \(T \subsetneq S^{*}\) omits an index with \(z_{j} - \tau^{*} > 0\), so
  \(\sum_{j \in T} (z_{j} - \tau^{*}) < \sum_{j \in S^{*}} (z_{j} - \tau^{*}) = 1\) and
  \(H_{\kappa} < \tau^{*}\).
  If \(c \le \kappa \le c_{\ge}\), then \(T = S^{*} \cup E\) with \(z_{j} = \tau^{*}\) for all
  \(j \in E\), so the sum equals \(1\) and \(H_{\kappa} = \tau^{*}\).
  If \(\kappa > c_{\ge}\), then \(T \supseteq S^{*}\); the indices of \(T \setminus S^{*}\) contribute
  terms \(z_{j} - \tau^{*} \le 0\), at least one of which is \(< 0\), so the sum is \(< 1\) and
  \(H_{\kappa} < \tau^{*}\).
 
  \emph{Subgradient.}
  By \eqref{equation:all-three-equal}, \(\tau^{*}(\bfz) = \max_{S \ne \emptyset} H(S)\) is a maximum
  of finitely many affine functions of \(\bfz\), hence convex.
  For any \(\bfz' \in \bbR^{K}\), writing \(H(S)(\bfz')\) for \(H(S)\) evaluated at \(\bfz'\),
  \[
    \tau^{*}(\bfz') \;\ge\; H(S^{*})(\bfz')
    \;=\; H(S^{*})(\bfz) + \tfrac{1}{c} \textstyle\sum_{j \in S^{*}} (z'_{j} - z_{j})
    \;=\; \tau^{*}(\bfz) + \bigl\langle \tfrac{1}{c} \mathbf{1}[\bfz > \tau^{*}],\, \bfz' - \bfz \bigr\rangle .
  \]
  Thus \(\tfrac{1}{c} \mathbf{1}[\bfz > \tau^{*}]\) is a subgradient of \(\tau^{*}\) at \(\bfz\), and
  adding the gradient of the affine term \(1 - z_{y}\) gives
  \eqref{equation:gradient-via-threshold}.
  Since \(S^{*}\) is the set of indices of the \(\kappabar = c\) largest entries, this is the vector
  in \eqref{equation:gradient-of-the-familywise-loss-main}.
\end{proof}
 
\begin{lemma}[Restriction]
  \label{lemma:restriction}
  Let \(t \le \tau^{*}\) and let \(U \subseteq \{1,\dots,K\}\) contain \(\{j : z_{j} > t\}\).
  Then \(U \ne \emptyset\), \(\{j : z_{j} > \tau^{*}\} \subseteq U\), and \(\tau^{*}\) is the unique
  root of \(g_{U}(s) := \sum_{j \in U} (z_{j} - s)_{+} = 1\).
  That is, the sub-vector \((z_{j})_{j \in U}\) has the same threshold and the same set of entries
  above it as \(\bfz\).
\end{lemma}
\begin{proof}
  \(z_{\max} > \tau^{*} \ge t\) by \Cref{lemma:root}, so an index of the maximum is in \(U\).
  \(\{z_{j} > \tau^{*}\} \subseteq \{z_{j} > t\} \subseteq U\).
  Every \(j \notin U\) has \(z_{j} \le t \le \tau^{*}\), hence contributes \(0\) to \(g(\tau^{*})\);
  therefore \(g_{U}(\tau^{*}) = g(\tau^{*}) = 1\).
  Uniqueness is \Cref{lemma:root} applied to the sub-vector.
\end{proof}
 
\begin{lemma}[The iteration \eqref{equation:newton-step}]
  \label{lemma:newton}
  Let \(t \le \tau^{*}\), \(A := \{j : z_{j} > t\}\) and \(t' := t + (g(t) - 1)/|A|\).
  Then \(A \ne \emptyset\), \(t' = H(A)\) and \(t \le t' \le \tau^{*}\).
  If \(\{j : z_{j} > t'\} = A\), then \(t' = \tau^{*}\).
  Moreover \(t_{0} := \max\{z_{\max} - 1, \frac{1}{K}(\sum_{j} z_{j} - 1)\} \le \tau^{*}\).
\end{lemma}
\begin{proof}
  \(A \ne \emptyset\) because \(z_{\max} > \tau^{*} \ge t\).
  Since \(g(t) = \sum_{j \in A} (z_{j} - t)\), we get
  \(t' = \frac{1}{|A|} (\sum_{j \in A} z_{j} - 1) = H(A)\), so \(t' \le \tau^{*}\) by
  \Cref{lemma:subset-lower-bound}; and \(g(t) \ge 1\) by \Cref{lemma:root}, so \(t' \ge t\).
  If \(\{z_{j} > t'\} = A\), then \(g(t') = \sum_{j \in A} (z_{j} - t') = |A| (H(A) - t') + 1 = 1\),
  so \(t' = \tau^{*}\) by uniqueness.
  Finally, \(z_{\max} - 1 = H(\{j_{0}\})\) and \(\frac{1}{K}(\sum_{j} z_{j} - 1) = H(\{1,\dots,K\})\),
  both \(\le \tau^{*}\) by \Cref{lemma:subset-lower-bound}.
\end{proof}
 
\begin{lemma}[Median-based search; proof of \Cref{theorem:linear-time-computation-of-the-gradient}]
  \label{lemma:median-pivot}
  Maintain a multiset \(U\) of undecided entries and the sum \(s\) and count \(c\) of the entries
  already classified as ``above''; initially \(U\) contains all \(K\) entries and \(s = c = 0\).
  While \(U \ne \emptyset\): let \(p\) be a median of \(U\); split \(U\) into
  \(G = \{u > p\}\), \(E = \{u = p\}\), \(L = \{u < p\}\); compute
  \(\mathrm{val} := s + \sum_{u \in G} u - (c + |G|)\, p\).
  If \(\mathrm{val} < 1\), add \(G \cup E\) to \((s, c)\) and set \(U \leftarrow L\); otherwise set
  \(U \leftarrow G\).
  At termination, \(c = |\{j : z_{j} > \tau^{*}\}| \ge 1\) and \((s - 1)/c = \tau^{*}\).
  With a worst-case linear-time median \citep{blum1973time}, the total time is \(\calO(K)\).
\end{lemma}
\begin{proof}
  We show by induction that before every iteration
  (I1) every entry classified ``above'' is \(> \tau^{*}\) and larger than every entry of \(U\), and
  (I2) every discarded entry is \(\le \tau^{*}\) and smaller than every entry of \(U\).
  Both hold vacuously at the start.
  Consider an iteration with pivot \(p \in U\).
  By (I1) the ``above'' entries exceed \(p\) and contribute \(s - c\,p\) to \(g(p)\);
  by (I2) the discarded entries are \(< p\) and contribute \(0\);
  \(G\) contributes \(\sum_{u \in G} (u - p)\) and \(E \cup L\) contribute \(0\).
  Hence \(\mathrm{val} = g(p)\).
  If \(g(p) < 1\), then \(p > \tau^{*}\) (\Cref{lemma:root}), so every entry of \(G \cup E\) is
  \(> \tau^{*}\), and it is larger than every entry of the new \(U = L\): (I1) is preserved, and (I2)
  is preserved because \(U\) shrank.
  If \(g(p) \ge 1\), then \(p \le \tau^{*}\), so every entry of \(E \cup L\) is \(\le \tau^{*}\) and
  smaller than every entry of the new \(U = G\): (I2) is preserved, and (I1) likewise.
  At termination every entry is classified, so by (I1) and (I2) the ``above'' entries are exactly
  those \(> \tau^{*}\). Then \(c \ge 1\) and \((s - 1)/c = H(\{j : z_{j} > \tau^{*}\}) = \tau^{*}\) by
  \Cref{theorem:threshold-identity}.
  Since \(p\) is a median, \(|G| \le |U|/2\) and \(|L| \le |U|/2\), and \(E \ne \emptyset\), so the new
  \(U\) has at most half the size. An iteration costs \(\calO(|U|)\), so the total is
  \(\calO(K + K/2 + K/4 + \cdots) = \calO(K)\).
  By \Cref{theorem:threshold-identity}, \((\tau^{*}, c)\) determine the loss and the subgradient
  \eqref{equation:gradient-via-threshold} with one more pass, which proves
  \Cref{theorem:linear-time-computation-of-the-gradient}.
\end{proof}
 
\subsection{Proof of \Cref{theorem:cpu-algorithm}}
\label{section:appendix:cpu-proof}
Let \(P\) denote the set of indices in the chunks processed so far, and let
\(\tau^{*}_{P}\) be the threshold of the sub-vector \((z_{j})_{j \in P}\) (\Cref{lemma:root}).
By \Cref{theorem:threshold-identity} applied to the sub-vector and to \(\bfz\),
\(\tau^{*}_{P} = \max_{\emptyset \ne S \subseteq P} H(S) \le \tau^{*}\).
\begin{lemma}[Invariants of \Cref{alg:threshold-search}]
  \label{lemma:chunk-invariants}
  After the first chunk step, at every point of \Cref{alg:threshold-search}:
  (a) \(t \le \tau^{*}_{P} \le \tau^{*}\);
  (b) \(t\) never decreases;
  (c) \(B \supseteq \{j \in P : z_{j} > t\}\), and \(B\) contains an index \(j\) with \(z_{j} > t\).
\end{lemma}
\begin{proof}
  \emph{Chunk step.}
  The new value of \(t\) is the maximum of the old one and \(z_{j_{I}} - 1 = H(\{j_{I}\})\), where
  \(j_{I}\) is an index of the chunk maximum and \(j_{I} \in P\) after the chunk is added to \(P\).
  So (b) holds, and (a) holds by \Cref{lemma:subset-lower-bound} applied to the sub-vector indexed
  by \(P\) (using (a) before the step and \(\tau^{*}_{P}\) non-decreasing in \(P\) for the old value).
  The indices of the chunk with \(z_{j} > t\) are appended; indices of earlier chunks that are not
  in \(B\) satisfied \(z_{j} \le t\) for an earlier, smaller value of \(t\).
  This gives the inclusion in (c).
  By \Cref{lemma:root} applied to the sub-vector, \(\tau^{*}_{P} < \max_{j \in P} z_{j}\), so by (a)
  an index of the maximum over \(P\) has \(z_{j} > t\) and therefore lies in \(B\).
 
  \emph{\textsc{Tighten}.}
  The first line removes from \(B\) only indices with \(z_{j} \le t\), so (c) is preserved and now
  \(B = \{j \in P : z_{j} > t\}\).
  In a round, \(\sum_{j \in B} (z_{j} - t) = \sum_{j \in P} (z_{j} - t)_{+} \ge 1\) by (a) and
  \Cref{lemma:root} applied to the sub-vector; hence \(t' \ge t\), which is (b).
  Moreover \(t' = H(B)\), so \(t' \le \tau^{*}_{P}\) by \Cref{lemma:subset-lower-bound}, which is (a).
  The set \(B' = \{j \in B : z_{j} > t'\}\) equals \(\{j \in P : z_{j} > t'\}\) because the indices
  of \(P\) outside \(B\) have \(z_{j} \le t \le t'\); it is nonempty as in the chunk step.
  So (c) holds for \((t', B')\), and for \((t', B)\) when \(B' = B\).
\end{proof}
 
\begin{proof}[Proof of \Cref{theorem:cpu-algorithm}]
  \emph{Correctness.}
  After the last chunk, \(P = \{1, \dots, K\}\) and \(\tau^{*}_{P} = \tau^{*}\).
  In the final call of \textsc{Tighten}, after its first line, \(B = \{j : z_{j} > t\}\) with
  \(t \le \tau^{*}\) (\Cref{lemma:chunk-invariants}), and every round is the step of
  \Cref{lemma:newton}: \(t' = H(\{j : z_{j} > t\})\) and \(B' = \{j : z_{j} > t'\}\).
  If \(B' = B\), then \(t' = \tau^{*}\) by \Cref{lemma:newton} and
  \(B = \{j : z_{j} > \tau^{*}\}\).
  Otherwise \(B' \subsetneq B\), so there are at most \(K\) rounds.
  If the budget of \(4K\) visited entries is exceeded first, then at that point \(t \le \tau^{*}\)
  and \(B \supseteq \{j : z_{j} > t\}\), so by \Cref{lemma:restriction} the sub-vector indexed by
  \(B\) has threshold \(\tau^{*}\) and the same entries above it, and \Cref{lemma:median-pivot}
  applied to it returns them.
  In both cases the last three lines output the loss and the subgradient of
  \Cref{theorem:threshold-identity}.
 
  \emph{Complexity.}
  The chunk steps cost \(\calO(K)\) in total.
  A call \(\textsc{Tighten}(t, B, 4)\) costs at most five passes over \(B\).
  It is made only when \(|B| \ge 4 b + 256\), where \(b\) is the size of \(B\) after the previous
  call (\(b = 0\) for the first call); since \textsc{Tighten} is the only operation that removes
  indices, at least \(\frac{3}{4}|B|\) of the current indices were appended after the previous call.
  Every index is appended once, so the sizes of \(B\) at the calls sum to at most
  \(\frac{4}{3} K\), and all these calls together cost \(\calO(K)\).
  The final call visits \(\calO(K)\) entries before it stops or hands over to the
  median-based search, which costs \(\calO(|B|)\) by \Cref{lemma:median-pivot}.
  Writing the gradient costs \(\calO(K)\).
\end{proof}
 
\begin{remark}[Floating point]
  The statements above are in exact arithmetic.
  The implementation enforces the following regardless of rounding: \(t\) never decreases (a round
  with computed \(t' \le t\) is treated as the fixed point), and the returned support is the
  content of \(B\), each index of which was compared with the returned threshold and found above
  it, while each discarded index was found at or below a threshold that is not larger.
  Hence the returned gradient always has exactly \(\kappabar \ge 1\) entries equal to
  \(1/\kappabar\) before the target is subtracted, and they are the logits above the returned
  threshold.
  If \(B\) becomes empty, which requires \(z_{(1)} - 1\) to round to \(z_{(1)}\), the sample is
  recomputed by the search of \Cref{lemma:median-pivot} in the shifted coordinates
  \(u_{j} = z_{j} - z_{(1)}\), using as mask threshold the largest discarded entry.
  We tested the kernel on \(14{,}960\) single-precision rows under \(7\)
  kernel configurations, for each of the AVX-512, AVX2 and scalar code paths, including logits of
  magnitude up to \(10^{30}\), rows with large common offsets, ties and constant rows, against an
  independent reference implementation; no row violated the property or the loss tolerance.
\end{remark}

\subsection{Implementation details of the CPU kernel}
\label{section:appendix:cpu-details}
\paragraph{Vectorization.}
The two operations per chunk, a maximum and a ``keep the entries above \(t\)'', and the two
operations of \eqref{equation:newton-step}, a masked sum and the same filter, take a few SIMD
instructions per 16 logits (AVX-512 compress; with AVX2, a permutation table).
 
\paragraph{Sparse backward pass.}
The gradient \eqref{equation:gradient-via-threshold} has \(\kappabar + 1\) non-zero entries, and
the forward pass already holds their indices in \(B\); we write only those.
For large gradient buffers (at least \(32\)\,MB, or at least \(8\)\,MB when the first sample
indicates a very sparse gradient) we request the memory from the operating system as zero pages,
so pages that contain no non-zero entry are never written.
 
\paragraph{Integration and arithmetic.}
The kernel is a C++ extension of PyTorch with its own autograd node: the forward pass computes the
loss and the gradient, and the backward pass returns the stored gradient.
Samples of a mini-batch are processed in parallel on PyTorch's thread pool.
Sums are accumulated in double precision as \(\sum_{j}(z_{j} - t)\), and \(t\) is only ever
increased, so that the returned support is exactly the set of logits above the returned threshold.
If rounding makes the search fail (this requires logits so large that \(z_{(1)} - 1\) is not
representable), the sample is recomputed by the median-based search in coordinates shifted by the
maximum logit, where \(\tau^{*} - z_{(1)} \in [-1, 0)\) (\Cref{lemma:root}).
 
\paragraph{Rows with nearly equal logits.}
\Cref{alg:threshold-search} starts from \(t = -\infty\), but any lower bound on \(\tau^{*}\) is a
valid starting value.
For rows whose logits are close together (at least \(30\%\) of a strided sample of \(16\) to
\(128\) logits lie within \(1\) of the sample maximum) and that fit in the second-level cache
(\(K \le 65{,}536\)), we compute the threshold of the sample, and one vectorized pass over the row
turns it into \(\frac{1}{|S|}(\sum_{j \in S} z_{j} - 1)\) for the set \(S\) of logits above it,
which is \(\le \tau^{*}\) by \Cref{lemma:subset-lower-bound}; the search starts from this value.
If, in addition, \(K < 512\) or the sample indicates that more than \(8\%\) of the row lies above
\(\tau^{*}\), the buffer \(B\) is not built: \eqref{equation:newton-step} is iterated on the whole
row, and the gradient is written by one vectorized comparison with \(\tau^{*}\).
The sample only selects a starting point; correctness rests on
\Cref{lemma:subset-lower-bound,lemma:newton} as before.

\section{Approximation of \(\kappabar\) for Gaussian logits}
\label{section:approximating-the-sparsity}
This appendix supports the paragraph ``Sparsity of the gradient'' of the main text: it states the
approximation of \(\kappabar\), compares it with simulations, and derives it.
 
\begin{theorem}[Roughly Stated]\label{theorem:approximation-of-kappabar}
  Suppose that the entries of \(\bfz \in \bbR^{K}\) are i.i.d.\
  \(\mathcal{N}(\mu,(\sigma/K)^{2})\).
  Then, for \(\sigma\) sufficiently large, as \(K \to \infty\),
  \[
    \kappabar/K \;\approx\; (1/\sigma)\sqrt{2\log(\sigma)} .
  \]
\end{theorem}

\paragraph{Plug-in estimate.}
For a given logit vector we use \(\hat\sigma := K \cdot \mathrm{std}(\bfz)\), where
\(\mathrm{std}\) is the empirical standard deviation, and estimate \(\kappabar/K\) by
\((C/\hat\sigma)\sqrt{2\log(\hat\sigma)}\) with a constant \(C \ge 1\).
 
\paragraph{On the Gaussian assumption.}
The assumption is made for the analysis only.
It is common in the literature: high-dimensional random feature maps behave like Gaussian random
variables in the thermodynamic limit \citep{goldt2022gaussian}, and the logits of large language
models are well described by a Gaussian with outliers \citep[Figure 1]{tang2024top}.
 
\begin{figure}[ht]
  \centering
  \includegraphics[width=0.48\textwidth]{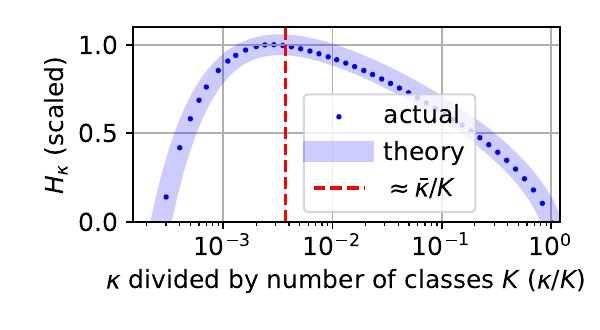}
  ~
  \includegraphics[width=0.48\textwidth]{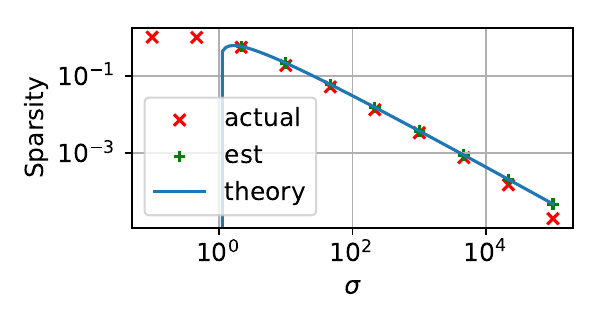}
  \caption{\textbf{(Left)} Simulated logits \(\bfz \sim \mathcal{N}(0, 0.1)\) with \(K = 10000\)
    (thus \(\sigma = 1000\) in the notation of \Cref{theorem:approximation-of-kappabar}).
    For each blue dot, the x-coordinate is \(\kappa/K\) and the y-coordinate is \(H_{\kappa}\)
    (subsampled to avoid clutter). The thick blue ribbon is the limiting curve as \(K \to \infty\).
  \textbf{(Right)} \(\kappabar\) versus \(\sigma\). The blue line is the prediction of
  \Cref{theorem:approximation-of-kappabar} with the actual \(\sigma\).}
    \label{figure:approximation-of-kappabar-vs-theory}
\end{figure}

\paragraph{Numerical check.}
\Cref{figure:approximation-of-kappabar-vs-theory} compares the approximation with simulated
logits for \(K = 10{,}000\): the points \((\kappa/K, H_{\kappa})\) follow the limiting curve derived
in \Cref{lemma:appendix:limiting-function} (left), and the predicted and the actual \(\kappabar\)
agree over several orders of magnitude of \(\sigma\) (right).
Since \((1/\sigma)\sqrt{2\log(\sigma)} \to 0\) as \(\sigma \to \infty\), the approximation predicts
that the gradient becomes sparser as the logit scale grows, which is what gradient descent on
separable data produces \citep[Lemma 1]{soudry2018implicit}.
 
\subsection{Setup of the derivation}
\label{section:appendix:estimating-kappa-bar}

Let \(\bfz \in \bbR^K\) be inputs to the familywise loss
and suppose that entries \(z_j\) of
\(\bfz\) are distributed according to the normal distribution \(\mathcal{N}(0, \sigma_{K}^{2})\) such that \(\sigma_{K} >0\) that linearly decays with respect to \(K\).
More precisely, we assume there exists a base quantity \(\sigma_{0} >0\) such that
\(\sigma_{K} = \sigma_{0} / K\) and
\[
z_j \overset{\mathrm{iid}}{\sim} \mathcal{N}(0,\sigma_{K}), \quad j = 1,\dots,K.
\]
Recall that the indices \((1),(2),\dots\) is the argsort of \(\bfz\) in descending order, i.e.,
\[
z_{(1)} \ge z_{(2)} \ge \cdots \ge z_{(K)}.
\]
Let \(\kappa \in \{1,\dots, K\}\) denote an arbitrary index.
Our goal is to approximate the \(\kappa\) that maximizes the \(\kappa\)-\shancs as \(K\to \infty\), i.e.,
\[
  \kappabar :=
  \underset{\kappa = 1,\dots, K}{\operatorname{argmax}} \,\,
- \frac{1}{\kappa}
+
\frac{1}{\kappa} \sum_{j=1}^{\kappa} z_{(j)}.
\]
To this end, we view
the \(\kappa\)-th \shancs as a function of \(\kappa/K\). Intuitively, the function can be visualized by plotting
\(
    \frac{ \kappa }{K}
\)
on the X-axis
and
the \(\kappa\)-th \shancs
\(
    - \frac{1}{\kappa}
    +
    \frac{1}{\kappa} \sum_{j=1}^{\kappa} z_{(j)}
\)
on the Y-axis. In other words, we plot the following scatter points
\[
  \left(
    \frac{ \kappa }{K}
    ,\,\,
    - \frac{1}{\kappa}
    +
    \frac{1}{\kappa} \sum_{j=1}^{\kappa} z_{(j)}
  \right),
  \qquad \kappa = 1,\dots, K.
\]
See
    \Cref{figure:approximation-of-kappabar-vs-theory}-left panel blue scatter points for the resulting plot.
    As \(K\to\infty\), this scatter plot converges to a function over the unit interval \([0,1]\) shown by the light blue colored thick band in \Cref{figure:approximation-of-kappabar-vs-theory}-left panel.
    Below, we will
    \begin{enumerate}
      \item calculate this limiting function,
      \item show that the limiting function attains exactly one argmax for any \(\sigma_{0}\), and
      \item derive an approximation of the argmax of the limiting function.
    \end{enumerate}
    If \(\kappa\) is constant as \(K \to \infty\), then \(\kappa/K \to 0\).
    Thus, we perform the calculation assuming that \(\kappa/K \to p \in (0,1)\) for some fixed value \(p\) as \(K \to \infty\).

    \subsection{Calculating the limiting function}
    We begin with a simpler calculation of the limiting curve as \(K\to \infty\) of
    \begin{equation}
      \label{equation:appendix:finite-inverse-quantile-function}
  \left(
    \frac{ \kappa }{K}
    ,\,\,
    \frac{z_{(\kappa)}}{\sigma_K}
  \right),
  \qquad \kappa = 1,\dots, K.
\end{equation}
Let \(\Phi\) be the cumulative distribution function (CDF) of the standard Gaussian/normal distribution (with zero mean and unit variance, denoted by \(\mathcal{N}(0,1)\)).
The \emph{quantile function}, denoted \(\Phi^{-1}\), is the (function-theoretic) inverse of \(\Phi\).

By assumption, we have that the \(z_{j}/\sigma_K\), \(j=1\dots, K\), are distributed according to \(\mathcal{N}(0,1)\).
Moreover,
\(z_{(\kappa)}/\sigma_K\) converges to the \((1-p)\)-th quantile as \(K \to \infty\). Therefore, we have that \(\lim_{K\to \infty}z_{(\kappa)}/\sigma_K = \Phi^{-1}(1-p)\) and
the graph
defined by
      \Cref{equation:appendix:finite-inverse-quantile-function}
      converges as \(K\to \infty\) to
      \[
(p, \Phi^{-1}(1-p)).
      \]

      Apply the ``running average'' operator to
      \Cref{equation:appendix:finite-inverse-quantile-function}, we have
\begin{equation*}
      % \label{equation:appendix:finite-running-average}
  \left(
    \frac{ \kappa }{K}
    ,\,\,
    \frac{1}{\kappa} \sum_{j=1}^{\kappa}\frac{z_{(j)}}{\sigma_K}
  \right),
  \qquad \kappa = 1,\dots, K.
\end{equation*}
Multiplying and dividing by \(K\), we have
\[
    \frac{1}{\kappa} \sum_{j=1}^{\kappa}\frac{z_{(j)}}{\sigma_K}
    =
    \frac{K}{\kappa} \sum_{j=1}^{\kappa}\frac{z_{(j)}}{\sigma_K} \cdot \frac{1}{K}
    \quad
    \to
    \quad
    \frac{1}{p} \int_{0}^{p} \Phi^{-1}(1-t) dt
\]
Since \(\sigma_{K} := \sigma_{0}/K\) by definition, we have
\[
  K \left(\frac{1}{\kappa} \sum_{j=1}^{\kappa}
  {z_{(j)}}\right)
  % \frac{z_{(j)}}{\sigma_K}
    \quad
    \to
    \quad
    \frac{
    \sigma_{0}
    }{p} \int_{0}^{p} \Phi^{-1}(1-t) dt.
\]
Moreover, using the fact that \(\kappa/K \to p\), we have
\[
  K \left(-\frac{1}{\kappa}+\frac{1}{\kappa} \sum_{j=1}^{\kappa}
  {z_{(j)}}\right)
  % \frac{z_{(j)}}{\sigma_K}
    \quad
    \to
    \quad
    -
    \frac{1}{p}
    +
    \frac{
    \sigma_{0}
    }{p} \int_{0}^{p} \Phi^{-1}(1-t) dt.
\]
The integral term on the right hand side is known as the expected shortfall of the normal/Gaussian distribution \cite{khokhlov2016conditional} and can be expressed as
\begin{equation}
  \label{equation:expected-shortfall-and-integral-of-gaussian-quantile}
\frac{1}{p}\int_{0}^{p} F^{-1}(1-t) dt
=
\frac{\varphi ( \Phi^{-1}(p) )}{p}
=:
\mathrm{ES}(p).
\end{equation}
where \(\varphi\) is the density function of the standard normal/Gaussian distribution.
Thus, we proved
the following Lemma:
\begin{lemma}\label{lemma:appendix:limiting-function}
  If the entries of \(\bfz \in \bbR^{K}\) are distributed iid according to \(\mathcal{N}(0,\sigma_{K})\) where \(\sigma_{K} := \sigma_{0}/K\). Suppose that
\(\kappa\) and \(K\) both tend to \(\infty\) such that \(\kappa/K \to p \in (0,1)\) where \(p\) is fixed.
Then
  \[
  K \left(-\frac{1}{\kappa}+\frac{1}{\kappa} \sum_{j=1}^{\kappa}
  {z_{(j)}}\right)
\quad
\to
\quad
    -
    \frac{1}{p}
    +
    \sigma_{0}
    \mathrm{ES}(p)
    \quad \mbox{as}
    \quad \kappa, K \to \infty.
  \]
\end{lemma}

%% convexity

\subsection{The limiting function has exactly one argmax}\label{sec:appendix:exactly-one-argmax}
The goal of this subsection is to prove the following \textbf{claim}: For any \(\sigma_{0} >0\), the function \(g(p) := \sigma_{0}\mathrm{ES}(p) - (1/p)\) has exactly one maximizer on \((0,1)\).
\begin{proof}[Proof of claim]
We will show that \(g'\), the first derivative of \(g\), is initially strictly increasing, then becomes strictly decreasing.
To simplify notation, let \(c = \sigma_{0}\).
For convenience, let $w = \Phi^{-1}(p)$.
By definition, we have
% \[g(p) = \frac{c\varphi(\Phi^{-1}(p)) - 1}{p}\]
\[g(p) = \frac{c\varphi(w) - 1}{p}
\]

By the quotient rule, we have
\[g'(p) = \frac{p \cdot \frac{d}{dp}(c\varphi(w) - 1) - (c \varphi(w) - 1) }{p^2}\]

By the chain rule:
\begin{equation}
  \label{equation:appendix:useful-formula-for-derivative-with-ES}
\frac{d}{dp}(\varphi(w)) = \varphi'(w) \cdot \frac{dw}{dp} = \varphi'(w)/\varphi(w) = -w
\end{equation}
where for the last identiy, we used the fact that for the standard normal distribution PDF \(\varphi\) we have
$-\varphi'(w) = -w\varphi(w)$ for any \(w \in \bbR\).
Therefore
\begin{equation}
  \label{equation:appendix:formula-for-the-derivative-of-the-ES}
g'(p) =
\frac{p \cdot (-cw) - (c\varphi(w) - 1)}{p^2} =
\frac{-pc\Phi^{-1}(p) - c\varphi(\Phi^{-1}(p)) + 1}{p^2}.
\end{equation}

Now, we claim that
\(
h(p):=
-pc\Phi^{-1}(p) - c\varphi(\Phi^{-1}(p)) + 1
\), i.e., the numerator of \(g'\),
is
\begin{enumerate}
  \item strictly decreasing on \((0,1)\),
  \item \(\lim_{p \to 0}h(p) =1\), and
  \item \(\lim_{p \to 1}h(p) =-\infty\).
\end{enumerate}
This will prove the desired claim.
Now, to see item 1, we show that \(h'\) is strictly negative:
\[
\frac{d}{dp}(p \cdot (-cw) - (c\varphi(w) - 1))
=
-cw -
pc
\frac{dw}{dp}
-
c
\frac{d}{dp}  \varphi(w).
\]
Using
\Cref{equation:appendix:useful-formula-for-derivative-with-ES}, we get
\(
\frac{d}{dp}  \varphi(w) = -w
\)
and so the above first and last term cancels out. Thus,
\[
\frac{d}{dp}(p \cdot (-cw) - (c\varphi(w) - 1))
=
-pc
\frac{dw}{dp}
= -pc\frac{1}{\varphi(\Phi^{-1}(p))}
\]
by the inverse function theorem
and
the fact that
$\frac{dp}{dw} = \frac{d}{dw} \Phi(w)= \varphi(w)$.
Since \(\varphi(\cdot)>0\) is always positive, we have that proved that \(h'\) is strictly negative and thus \(h\) is strictly decreasing on \((0,1)\).

For item 3, clearly, as \(p \to 1\), we have that \(h(p) \to -\infty\).
This is because
\(\lim_{p\to \{0,1\}} c \varphi(\Phi^{-1}(p)) =c \varphi(\pm\infty) = 0\).
Moreover,
\(\lim_{p\to 1}-pc\Phi^{-1}(p)
=-c
\lim_{p\to 1}\Phi^{-1}(p)
=-c
(+\infty)
=-\infty.
\)

It remains to prove that, \(\lim_{p\to 0} h(p) = 1\).
To see this, note that we already showed
\(\lim_{p\to 0} c \varphi(\Phi^{-1}(p)) = 0\).
Below, we show that \(
\lim_{p\to 0} -p  \Phi^{-1}(p) =0
\).
It is well known that \begin{equation}
  \label{equation:appendix:sqrt-approx-of-quantile}
\Phi^{-1}(p) \sim -\sqrt{-2\ln(p)}
\end{equation} for \(p\) near \(0\).
See paragraph immediately following \citep[Eqn.~(18)]{machler2022asymptotic}.
Thus, \(\lim_{p\to 0} p \Phi^{-1}(p) =-\sqrt{-2\lim_{p\to 0}  p^{2} \ln(p)}\).
Now, by L'H\^{o}pital's rule, we have
\[
  \lim_{p\to 0}  p^{2} \ln(p)
  =
  \lim_{p\to 0}  \frac{ \ln(p)}{1/p^{2}}
  =
  \lim_{p\to 0}
  \frac{1/p}{-2/p^{3}}
  =
  \lim_{p\to 0} -p^{2}/2 = 0
\]
as desired.
\end{proof}

\subsection{An approximation of the argmax}
\begin{lemma}\label{lemma:derivative-of-the-expected-shortfall}
  The derivative of \(\mathrm{ES}\), the expected shortfall of the standard normal/Gaussian distribution as defined in
  \Cref{equation:expected-shortfall-and-integral-of-gaussian-quantile},
  is given by
  \[
    \frac{d \mathrm{ES} }{d p}  (p)
    =
    \frac{
      \Phi^{-1} (1-p) p - \varphi(\Phi^{-1}(1-p))
    }{p^{2}}.
  \]
\end{lemma}
\begin{proof}
  By a similar calculation as before in deriving \Cref{equation:appendix:formula-for-the-derivative-of-the-ES}, we have that
  \[
    \frac{
      d
    \mathrm{ES}
    }{dp}
    (p) =
\frac{-p\Phi^{-1}(p) - \varphi(\Phi^{-1}(p)) }{p^2}.
  \]
  To conclude, we note that \(\Phi^{-1}(\bullet) = - \Phi^{-1}(1-\bullet)\) and moreover that \(\varphi(-\bullet) = \varphi(\bullet)\)
\end{proof}

Given the earlier
\textbf{claim} ``For any \(\sigma_{0} >0\), the function \(g(p) := \sigma_{0}\mathrm{ES}(p) - (1/p)\) has exactly one maximizer on \((0,1)\)'' and
the formula for the derivative in
\Cref{lemma:derivative-of-the-expected-shortfall}, we can now compute
\[
\underset{p \in (0,1)}{\operatorname{argmax}}\,\,\sigma_{0}\mathrm{ES}(p) - \frac{1}{p}.
\]
To this end, we take derivative w.r.t \(p\) of the above and solve for roots (which we know there is exactly one by the claim):
\[
  \sigma_{0}
    \frac{
      \Phi^{-1} (1-p) p - \varphi(\Phi^{-1}(1-p))
    }{p^{2}}
    +
    \frac{1}{p^{2}}
    =0
\]
Simplifying, we have
\begin{equation}
  \label{equation:appendix:original-relationship-between-p-and-sigma0}
      \Phi^{-1} (1-p) p - \varphi(\Phi^{-1}(1-p))
    =-\frac{1}{
  \sigma_{0}
  }.
\end{equation}
The goal is to solve for \(p\) in terms of \(\sigma_{0}\).
However, the normal CDF \(\Phi\) and its inverse \(\Phi^{-1}\) do not have closed-forms.
Thus, we resort to approximations.
Let \(t := \Phi^{-1}(1-p)\). Thus, \(p = 1-\Phi(t)\).
Then
\begin{equation}
  \label{equation:appendix:tp-reduction}
      \Phi^{-1} (1-p) p - \varphi(\Phi^{-1}(1-p))
      =
      tp - \varphi(t).
\end{equation}
Our goal now is to replace the LHS of \Cref{equation:appendix:tp-reduction} by a simpler expression purely defined in terms of \(p\) which holds when \(p\) is close to \(0\).
We recall
Eqn.~(2.38)
from
\cite{small2010expansions}
which says the following
\[
  \frac{\varphi(t)}{t} \left(
  1 - \frac{1}{t^{2}} + \frac{3}{t^{4}} - \frac{3\cdot 5}{t^{6}}
  +\frac{3\cdot 5 \cdot 7}{t^{8}} - \cdots
\right)
  \sim 1 - \Phi(t)
  \quad \mbox{as}
  \quad
t \to +\infty.
\]
We will only be using the first two terms of the series approximation:
\begin{equation}
  \label{equation:appendix:second-order-approximation-to-the-root-equation}
  \frac{\varphi(t)}{t} \left(
    1 - \frac{1}{t^{2}}
\right)
  \approx 1 - \Phi(t)
  \quad \mbox{as}
  \quad
t \to +\infty.
\end{equation}
Recall earlier that we let \(t := \Phi^{-1}(1-p)\) and thus \(p = 1-\Phi(t)\). From
\Cref{equation:appendix:second-order-approximation-to-the-root-equation} we get
\begin{equation}
  \label{equation:appendix:p-second-order-approximation-to-the-root-equation}
  \frac{\varphi(t)}{t} \left(
    1 - \frac{1}{t^{2}}
\right)
  \approx p
  \quad \mbox{as}
  \quad
t \to +\infty.
\end{equation}
Isolating  \(\varphi(t)\), we get
\begin{equation}
  \label{equation:appendix:iso-second-order-approximation-to-the-root-equation}
  \varphi(t)
  \approx
  pt
  \left(
  \frac{1}{
    1 - \frac{1}{t^{2}}
  }
\right  )
  \quad \mbox{as}
  \quad
t \to +\infty.
\end{equation}
Using the geometric series and truncating to the first two terms, the above RHS can be approximated by
\[
  pt
  \left(
  \frac{1}{
    1 - \frac{1}{t^{2}}
  }
\right  )
=
pt ( 1 + t^{-2} + t^{-4} + \cdots)
\approx
pt ( 1 + t^{-2} )
\]
from which we deduce that
\[
  \varphi(t) \approx pt  + \frac{p}{t}
\]
Now, \Cref{equation:appendix:tp-reduction}
simplifies as
\[      \Phi^{-1} (1-p) p - \varphi(\Phi^{-1}(1-p))
      =
      tp - \varphi(t)
      = - p/t
\]
when \(p\) is close to \(0\).
To finally eliminate \(t\), we recall \Cref{equation:appendix:sqrt-approx-of-quantile} from earlier
where
\(
\Phi^{-1}(p) \sim -\sqrt{-2\log(p)}
\)
Since \(t = \Phi^{-1}(1-p) = - \Phi^{-1}(p) \), we finally get
\[
  \Phi^{-1} (1-p) p - \varphi(\Phi^{-1}(1-p))
      =
      tp - \varphi(t)
      = p/ \sqrt{-2 \ln(p)}.
\]
Note that throughout, we are interested in the approximation when \(p\) is near \(0\). This precisely corresponds to the case when \(t \to +\infty\) in which case all approximations used above hold.
Thus, we have proven:

\begin{lemma}\label{lemma:appendix:approximation}
  When \(p\) is near \(0\), we have
  \[
      \Phi^{-1} (1-p) p - \varphi(\Phi^{-1}(1-p))
      \approx
      \frac{p}{\sqrt{-2 \ln(p)}}.
  \]
\end{lemma}
Recall that our original motivation is to solve for \(p\) in terms of \(\sigma_{0}\)
in
\Cref{equation:appendix:original-relationship-between-p-and-sigma0}.
Given the above lemma, we only need to solve for \(p\) in terms of \(\sigma_{0}\) in the following equation:
\begin{equation}
  \label{equation:solving-the-argmax-approximator}
\frac{p^{2}}{2 \ln(1/p)} = \frac{1}{\sigma_{0}^{2}}.
\end{equation}
\begin{lemma}\label{lemma:appendix:approximation2}
  An approximation to \eqref{equation:solving-the-argmax-approximator}
  is given by
  \(    p \approx \frac{1}{\sigma_{0}} \sqrt{2 \ln (\sigma_{0})}.
\)
\end{lemma}
Note that the above Lemma together with  immediately implies our desired result
\Cref{theorem:approximation-of-kappabar}.
\begin{proof}
For notational convenience, we write \(a = \frac{1}{\sigma_{0}}\).  So the goal is to solve for
 \(p\)
 in
 \[
\frac{p^{2}}{2 \ln(1/p)} = a^{2}.
 \]
 Rearranging, we see that the above is equivalent to
\[
\ln(p) = -\frac{p^{2}}{2a^{2}}.
 \]
 Letting \(y := -\ln(p)\), we have
 \[
   ye^{2y} = \frac{1}{2a^{2}}.
 \]
 Further substituting \(u = 2y\), we get
 \[
   ue^{u} = \frac{1}{a^{2}}.
 \]
 To solve the above, recall that the Lambert W function \citep[Eqn.~(1)]{corless1997sequence} satisfies the equation
 \[
   W(z)e^{W(z)} = z
 \]
 So if we substitute \(W(z) = u\), then we have
 \[
   W(z)e^{W(z)} = z = W^{-1}(u) = \frac{1}{a^{2}}.
 \]
 In other words, \(u = W(\frac{1}{a^{2}})\).
 Substituting back \(p\), we see that
 \begin{equation}
   \label{equation:x-in-terms-of-W-function}
   p = \exp\left(-\frac{1}{2} W\left(\frac{1}{a^{2}}\right)\right)
\end{equation}
It is well known
(e.g., see \cite[\S~4.2]{corless1997sequence})
that
 \[
   W(
   z
   )
   =\ln(z) - \ln (\ln(z)) +o(1).
 \]
 Plugging the above into \eqref{equation:x-in-terms-of-W-function}, we have
 \begin{align}
   p
   &= \exp\left[ -\tfrac{1}{2}
     \left(
     \ln\left(\tfrac{1}{a^{2}}\right)
     -\ln\left(
       \ln\left(\tfrac{1}{a^{2}}\right)
     \right     )
\right     )
   \right ]
   = \exp\left[
     \ln(a)
     +\ln\left(
       \sqrt{
         \ln\left(\tfrac{1}{a^{2}}\right)
       }
     \right     )
   \right ]
     \\
   &=
     a \sqrt{\ln(1/a^{2})}
     =
     a \sqrt{2\ln(1/a)}
\end{align}
Substituting in \(\sigma_{0} = 1/a\), we get the desired approximation.
\end{proof}

\subsection{Putting it all together}
In view of \Cref{lemma:appendix:limiting-function}
and
\Cref{sec:appendix:exactly-one-argmax}'s main claim, we
only have to solve for the (unique) root of
\[-
    \frac{1}{p}
    +
    \sigma_{0}
    \mathrm{ES}(p).
\]
By
\Cref{lemma:derivative-of-the-expected-shortfall}, this reduces to solving the
\Cref{equation:appendix:original-relationship-between-p-and-sigma0}.
Using the approximations in \Cref{lemma:appendix:approximation} and \Cref{lemma:appendix:approximation2}, we get the desired result.

\section{Additional experimental details}
\label{section:appendix:experiments}

This section reports the training experiments summarized in the main text.
In all of them the FW and the CE loss are compared under the same pipeline with a parameter-free
optimizer: Prodigy \citep{mishchenko2024prodigy}, GDTUO \citep{chandra2022gradient} or DoG
\citep{ivgi2023dog}.
Hardware, statistical tests and instructions for reproducing each experiment are in
\Cref{section:appendix:experiments}.

\subsection{Linear classifiers on UCI datasets}
\paragraph{Setup.}
We use the 120 UCI datasets curated by \cite{fernandez2014we}\footnote{The original count is 121
datasets. However, the \textsc{nursery} dataset had corrupted labels, which we exclude.}.
Linear models are trained with Prodigy at its default settings, without regularizers.
For each loss and each dataset we run five trials with random initializations.
We use the original training/testing sets and split the training set 80/20 to obtain a validation
set.
For each run we record the test accuracy at the iteration with the highest validation accuracy, and
aggregate the five runs by their median.
A dataset is counted as having a ``large'' feature space if it has at least \(30\) features.

\paragraph{Results.}
\Cref{table:UCI-dataset-performance} counts the datasets on which each loss attains the higher
aggregate accuracy.
With at least \(30\) features, FW wins more often (Wald confidence interval \((0.464, 0.990)\), not
statistically significant); with fewer features, CE wins more often (Wald confidence interval
\((0.540, 0.825)\)).
The tests are described in \Cref{section:appendix:UCI}.

\begin{table}[ht!]
\centering
\begin{tabular}{@{}lcc@{}}
\toprule
  \multirow{2}*{
  \begin{tabular}{l}
\(\#\)datasets where \\ \underline{\hspace{2em}} loss won
  \end{tabular}
  } & \multicolumn{2}{c}{\(\#\)features}
  \\
\cmidrule(lr){2-3}
                      & \hspace{-0.5em}$< 30$     & \hspace{-0.5em} $\geq 30$   \\
\midrule
FW                     & 22      & \textbf{15}    \\
CE                     & \textbf{44}      & 9    \\
Both                  & 20      & 10    \\
  \midrule
Total                  & 86      & 34    \\
\bottomrule
\end{tabular}
\vspace{1em}
\caption{Performance of familywise loss vs cross entropy loss-trained linear classifiers on the 90 datasets panel subset of the UCI dataset from \cite{arora2020harnessing}}
\label{table:UCI-dataset-performance}
\end{table}

\subsection{MNIST with GDTUO}
\paragraph{Setup.}
GDTUO (``Gradient Descent: The Ultimate Optimizer'', \citealp{chandra2022gradient}) learns the
learning-rate schedule by optimizing the learning rates on the fly with another iterative method.
We use the example recipe of its official PyTorch
implementation\footnote{https://github.com/kach/gradient-descent-the-ultimate-optimizer/}, a
two-layer fully-connected network on MNIST \citep{lecun1998gradient} trained for 5 epochs, and
change only the loss.
Each loss is run 10 times.

\paragraph{Results.}
\Cref{fig:mnist-gdtuo}-left shows test accuracy versus wall clock time.
The final median test accuracy is \(97.93\%\) with FW and \(97.66\%\) with CE.

\subsection{CIFAR-100 with DoG}
\paragraph{Setup.}
We train a ResNet18 on CIFAR-100 \citep{krizhevsky2009learning}.
GDTUO was numerically unstable in this setting, possibly because of the larger number of classes,
so we use DoG, which also selects the learning rate and its schedule automatically but by a
different approach.
Optimizer hyperparameters follow the recipe in the official DoG
repository\footnote{\texttt{example.py} from \url{https://github.com/formll/dog/}}.
Since that recipe is for MNIST, the remaining hyperparameters are common choices for CIFAR-100
(e.g., batch size 128 and 200 epochs).

\subsection{ImageNet using Prodigy}

For a larger scale experiment, we train ResNet18 on the ImageNet \citep{russakovsky2015imagenet} dataset. We observe that by using Prodigy (an hyperparameter-free optimizer) the difference in performance when using the FW and the CE loss is essentially indistinguishable.
However, we note that using hand-tuned hyperparameters, it is possible to reach $>$70\% (top-1) accuracy.

At the root of the \redact{TODO}{supplemental material folder}, run the following for GPU

\begin{lstlisting}
conda create -n fwradix python -y
conda activate fwradix
pip install torch
cd code
TORCH_CUDA_ARCH_LIST="8.9" python setup.py build_ext --inplace
cd benchmark
python CE_FW_perf.py
\end{lstlisting}

\subsection{Linear classifiers on UCI datasets}\label{section:appendix:UCI}

\textbf{Detailed metadata and numerical results}.
The full table of
metadata for the 121 UCI datasets
can be found
at
\redact{\href{https://github.com/YutongWangML/HANN/blob/main/UCI_classification/metadata/datasets_summary.csv}{[link to metadata for the UCI datasets]}.}{\texttt{UCI-metadata.csv} in the supplemental materials.}

The full table of test accuracy comparison can be found
at \redact{TODO}{\texttt{UCI-accuracy-table.csv} in the supplemental materials.}

\textbf{Statistical tests}.
The Wald test is implemented according to the procedure described in

\url{https://stats.stackexchange.com/a/530465}

for rejecting the hypothesis that ``FW-loss training does not beat CE-loss training''. Conceptually, we can view this hypothesis as testing if we should reject that ``a coin is not biased toward head''.
The code is in \texttt{UCI-experiments/Wald\_Test.py}

\textbf{Hardware}.
The experiments were run on a machine with an x86 architecture Intel Xeon w5-3435X CPU with 32 logical cores.
The full batch of experiments (including 5 replicates) completed in under 12 hours.
GPU is not used for this experiment.

\textbf{Reproducing the experiments}. Follow the instruction at the beginning of
\Cref{section:appendix:experiments}, then follow the instruction in \verb|UCI-experiments/README.md|
in \redact{TODO}{the supplemental materials.}

\subsection{MNIST using ``Gradient Descent: The Ultimate Optimizer'' (GDTUO)}\label{section:appendix:UCI-statistical-test}

\textbf{Reproducing the experiments}. Follow the instruction at the beginning of
\Cref{section:appendix:experiments}, then follow the instruction in \verb|MNIST/README.md|
in \redact{TODO}{the supplemental materials.}

\textbf{Hardware}.
The experiments were conducted on a NVIDIA RTX 5000 Ada on the same machine as in
\Cref{section:appendix:UCI}.
The full batch of experiments (including 10 replicates for each loss function) completed in under 3 hour.

\subsection{CIFAR-100}
\label{section:appendix:cifar-100}

\textbf{Additional replicates}. We plot the results from four additional replicates for both the FW and CE losses.
We use GPUs from different machines to compute these replicates so the wallclock time are not directly comparable. Thus
\Cref{fig:mnist-gdtuo} plots the epochs on the X-axis.

\textbf{Hardware}.
The experiments in the main article were conducted on a NVIDIA RTX 5000 Ada on the same machine as in
\Cref{section:appendix:UCI}.
The additional replicate experiments were  run on two cloud computing instances with
Quadro RTX 6000 GPUs.
The full batch of experiments (including four replicates for each loss function) completed in under 12 hour.

\textbf{Reproducing the experiments}. Follow the instruction at the beginning of
\Cref{section:appendix:experiments}, then follow the instruction in \verb|CIFAR-100/README.md|
in \redact{TODO}{the supplemental materials.}

\subsection{Other Distributions}
We additionally evaluate the CPU and GPU runtime under Laplace and
Gaussian-mixture logit distributions.
\label{section:appendix:distributions}
\begin{figure}[H]
  \centering
\includegraphics[width=\textwidth]{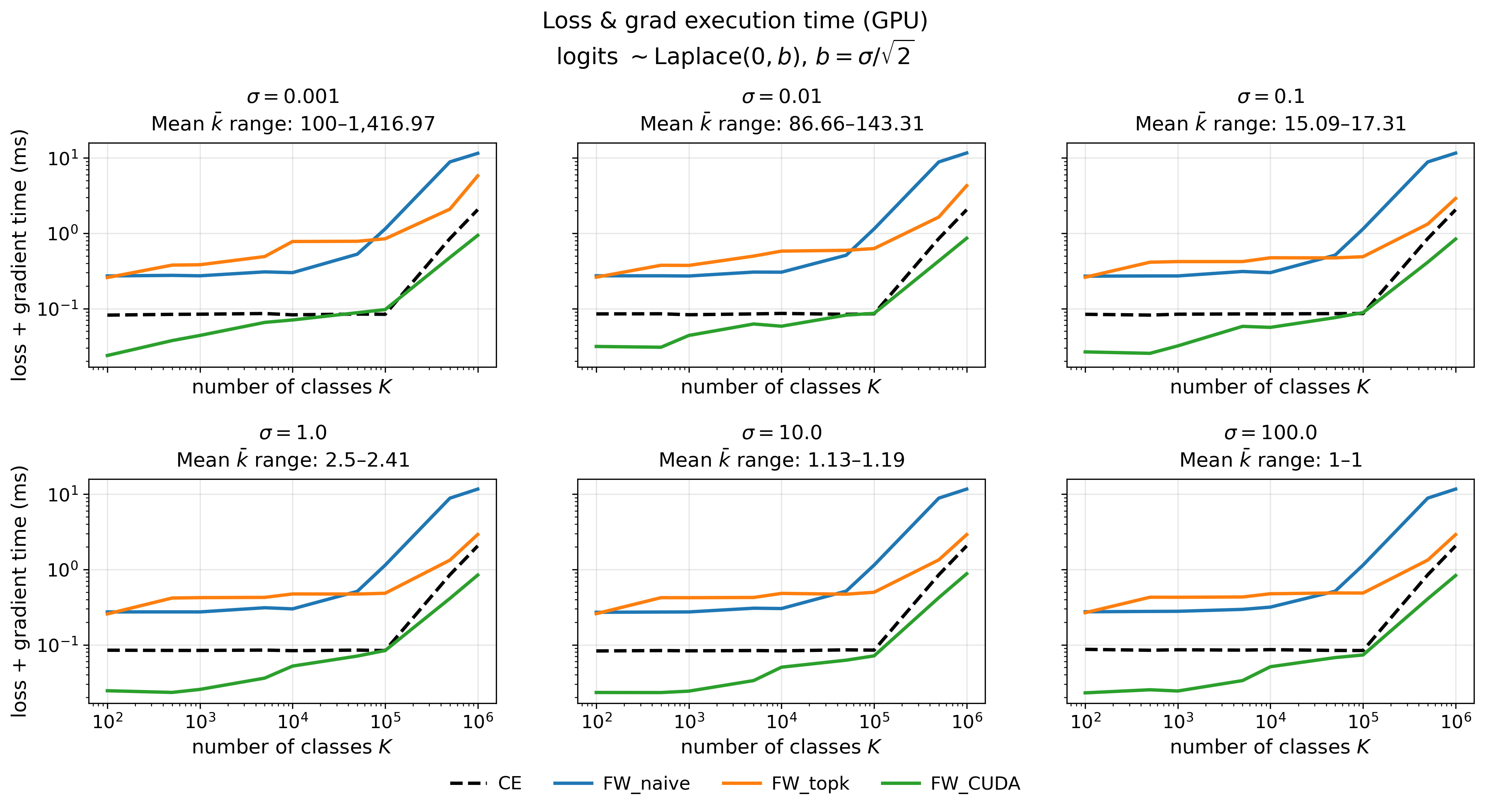}
  \caption{Laplace distribution}
  \label{fig:GPU-laplacian-time-comparison}
\end{figure}
\begin{figure}[H]
  \centering
\includegraphics[width=\textwidth]{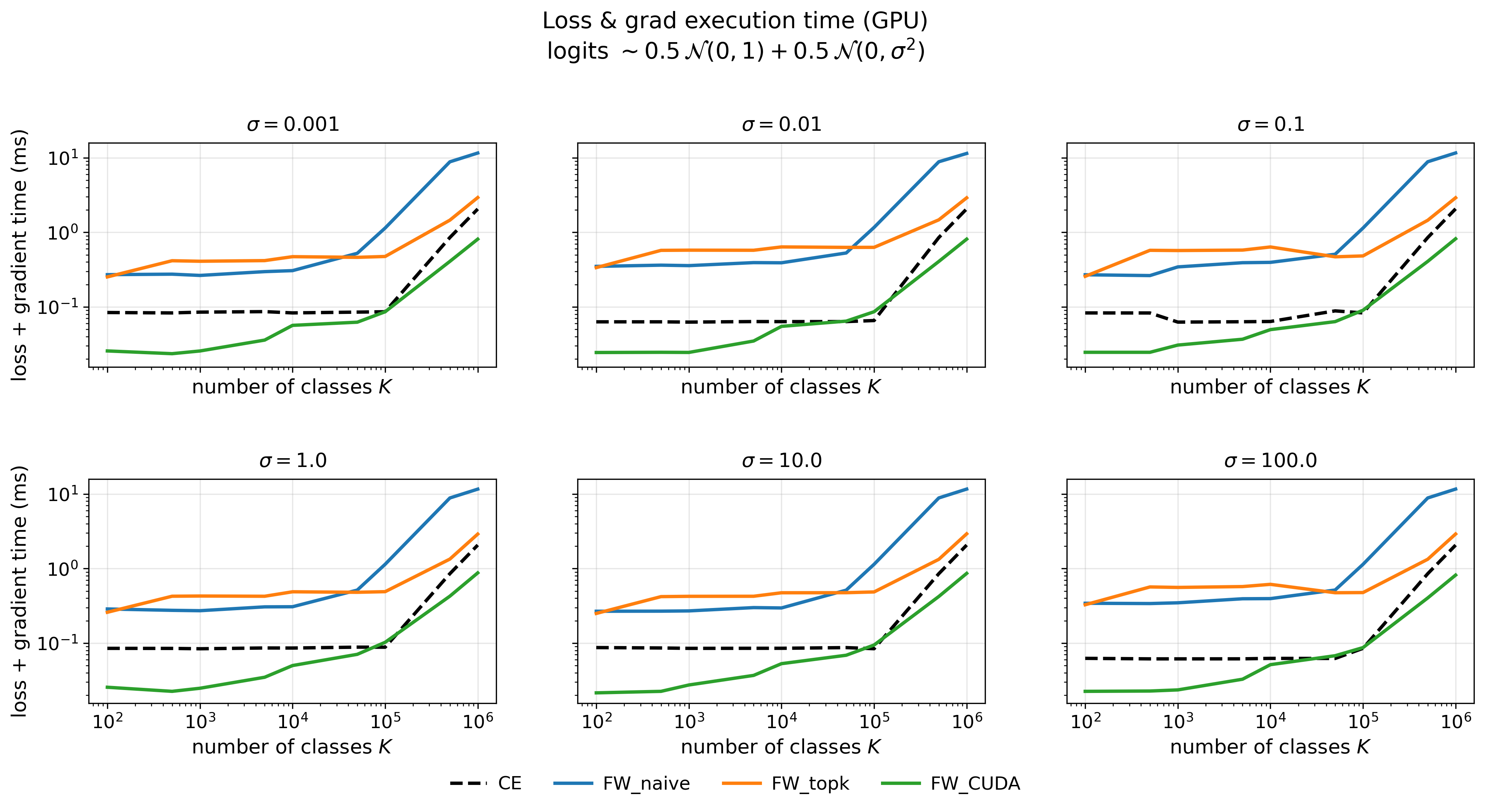}
 \caption{Mixed Gaussian distribution}

  \label{fig:GPU-mixed-time-comparison}
\end{figure}
\begin{figure}[ht]
  \centering
  \includegraphics[width=\textwidth]{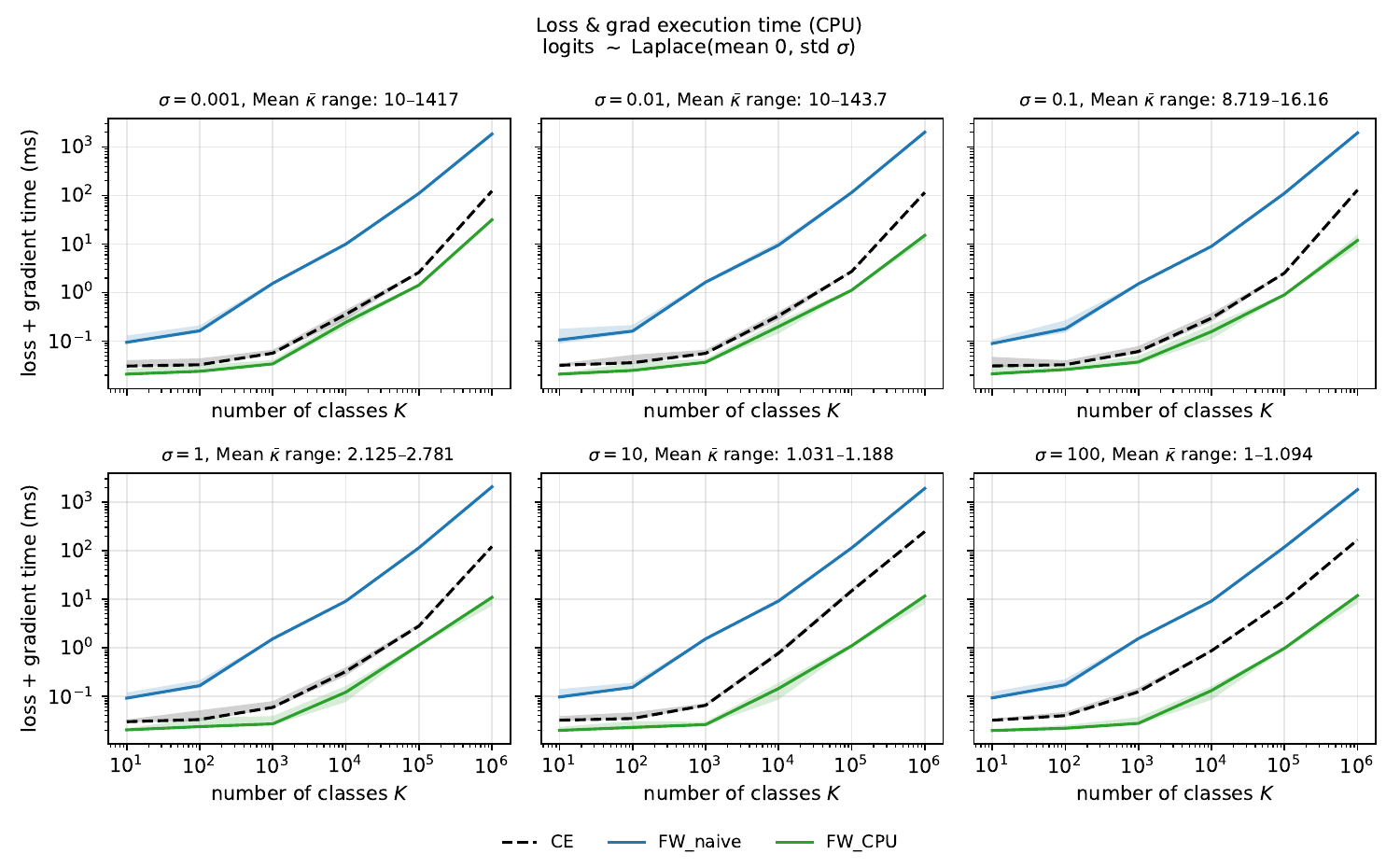}
  \caption{CPU loss and gradient computation time for Laplace logits with mean \(0\) and standard
  deviation \(\sigma\) (batch size 32, AVX-512, two threads). Lines are medians and bands are the
  20--80 quantiles.}
  \label{fig:appendix:cpu-laplace}
\end{figure}
\begin{figure}[ht]
  \centering
  \includegraphics[width=\textwidth]{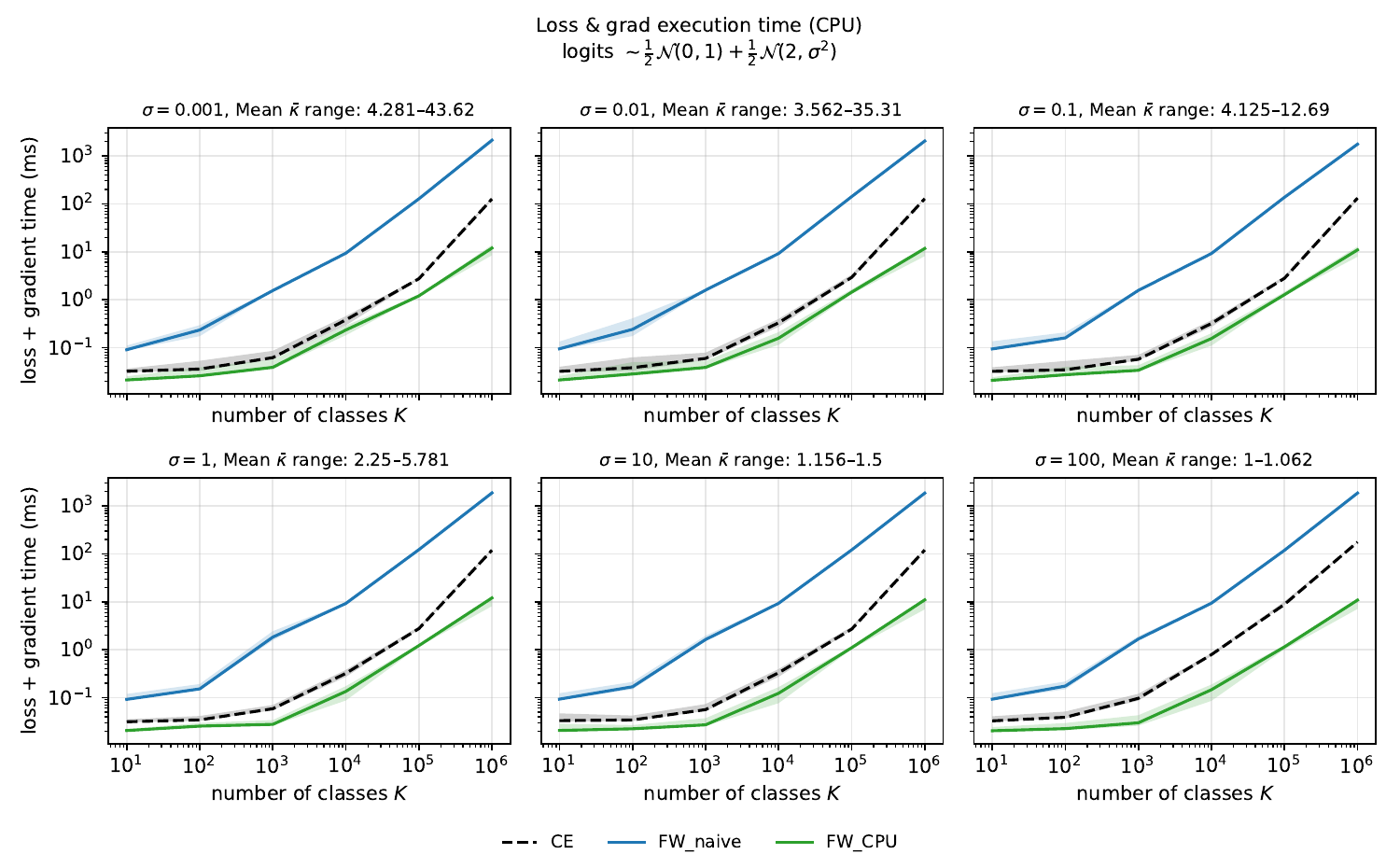}
  \caption{CPU loss and gradient computation time for mixture logits: each logit is drawn from
  \(\mathcal{N}(0,1)\) or from \(\mathcal{N}(2, \sigma^{2})\) with probability \(1/2\) each (batch size
  32, AVX-512, two threads). Lines are medians and bands are the 20--80 quantiles.}
  \label{fig:appendix:cpu-mixture}
\end{figure}
\subsubsection{Effect of $\sigma$} 
\label{section:appendix:sigma-effect} Both GPU and CPU experiments lead to different maximizing $\kappabar$  values. As $ \sigma$ increases, the mean maximizing $\kappabar$ decreases from hundreds or thousands for small $ \sigma$ to values close to one for large $ \sigma$. 
\textbf{GPU} - Despite this variation in $\kappabar$, radix FW shows almost the same runtime behavior across the six distributions. This suggests that the implementation is not very sensitive to the sparsity induced by the input distribution over the tested range. The improvement comes from filtering the candidate set before radix partitioning and restricting each subsequent pass to the single bin selected for refinement, thereby avoiding both full sorting and top-k selection.

\section{Naive implementation of the familywise loss}
\label{section:appendix:naive-impl-of-fwloss}

\begin{lstlisting}[language=python,numbers=none]
z = logits # has shape (batch_size, num_classes)
y = target # has shape (batch_size)
z_at_y = z[torch.arange(z.shape[0]), y]
z_hnscs = torch.sort(z, axis=-1, descending=True).values
z_hnscs.cumsum_(axis=-1)
z_hnscs.sub_(1)
kappa_inv = 1/torch.arange(1,num_classes+1).to(device)
z_hnscs.mul_(kappa_inv[torch.newaxis,:])
losses = 1 - z_at_y + torch.max(z_hnscs, axis=-1).values
\end{lstlisting}

% \begin{figure}
%   \centering
%   \vspace{-2em}
%   \includegraphics[width=0.42\textwidth]{notebooks/02_speed_test_synth/cuda_results/01_time_comparison_cuda_Ks.pdf}
%   ~
%   \includegraphics[width=0.42\textwidth]{notebooks/02_speed_test_synth/cuda_results/01_time_comparison_cuda_sigmas.pdf}
%   \caption{
%     Similar to
%   \Cref{fig:time-comparison-cpu} but on GPU.
%   A loss-and-grad calculation speed comparison between the naive implementation of the FW loss (\Cref{section:appendix:naive-impl-of-fwloss}), topk-based implementations
% (See \Cref{sec:pure-pytorch})
%   , and heaps based  implementation, and the CE loss of PyTorch.
%     Each run is repeated 20 times. The 20-80 quantiles are plotted as fill-betweens.
%     \textbf{(Left)} Execution time versus the number of classes.
%     \textbf{(Right)} Execution time versus \(\sigma\) where \(\sigma\) is proportional to the standard deviation of the logits as in \Cref{theorem:approximation-of-kappabar}.
%   }
%   \label{fig:time-comparison-gpu}
%     % \vspace{-2em}
% \end{figure}

\end{document}